\UseRawInputEncoding
\pdfoutput=1

\documentclass[11pt]{article}

\usepackage[preprint]{acl}
\usepackage{fix-cm}
\usepackage{times}
\usepackage{microtype}
\usepackage{graphicx}   
\usepackage{caption}    
\usepackage{multirow}
\usepackage{tabularx}
\usepackage{booktabs}
\definecolor{citecolor}{HTML}{0071BC}
\definecolor{linkcolor}{HTML}{ED1C24}
\definecolor{LGray}{gray}{0.97}
\usepackage{multicol}
\usepackage{colortbl}
\usepackage{xcolor}
\definecolor{tabhighlight}{HTML}{e5e5e5}
\usepackage{color}
\usepackage{xspace}
\usepackage{pifont}
\usepackage{colortbl} 
\usepackage{subcaption}

\usepackage{tcolorbox}
\tcbuselibrary{listings}
\usepackage{listings}
\usepackage{makecell}

\usepackage{graphicx}
\usepackage{tikz}
\usetikzlibrary{trees}

\definecolor{darkgreen}{rgb}{0.0, 0.5, 0.0}  
\newcommand{\cmark}{\textcolor{darkgreen}{\scalebox{1}[1.0]{\ding{51}}}}
\newcommand{\xmark}{\textcolor{red}{\ding{55}}}  

\lstdefinelanguage{json}{
    basicstyle=\ttfamily\scriptsize,
    numbers=left,
    numberstyle=\tiny,
    stepnumber=1,
    breaklines=true,
    showstringspaces=false,
    frame=single,
    backgroundcolor=\color{gray!10},
    keywordstyle=\color{blue},
    stringstyle=\color{red}
}

\usepackage[T1]{fontenc}

\usepackage[utf8]{inputenc}

\usepackage{inconsolata}

\usepackage{graphicx}

\title{
    \begin{minipage}{0.12\textwidth} 
        \raggedleft
        \includegraphics[height=1.5cm]{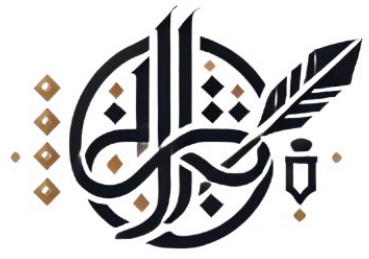} 
    \end{minipage}%
    \hspace{0.005cm} 
    \begin{minipage}{0.8\textwidth} 
        \centering
        \textbf{Fann or Flop: A Multigenre, Multiera Benchmark for Arabic Poetry Understanding in LLMs}
    \end{minipage}
}

\author{\\ {Wafa Alghallabi}\textsuperscript{1$\dagger$} \quad{Ritesh Thawkar}\textsuperscript{1$\dagger$} \quad{Sara Ghaboura}\textsuperscript{1$\dagger$}\quad {Ketan More}\textsuperscript{1$\dagger$} \quad{Omkar Thawakar}\textsuperscript{1$\dagger$} \\
    {Hisham Cholakkal}\textsuperscript{1} \quad  
     {Salman Khan}\textsuperscript{1,2} \quad  {Rao Muhammad Anwer}\textsuperscript{1,3}\\
     \fontsize{11pt}{12pt}\selectfont \textsuperscript{1}Mohamed bin Zayed University of AI,  \textsuperscript{2}Australian National University,
     \textsuperscript{3}Aalto University \\
     \fontsize{10pt}{12pt}\selectfont \{{wafa.alghallabi, sara.ghaboura, omkar.thawakar}\}@mbzuai.ac.ae \\
 {\hypersetup{urlcolor=blue}
\fontsize{11pt}{12pt}\selectfont \href{https://mbzuai-oryx.github.io/FannOrFlop/}{https://mbzuai-oryx.github.io/FannOrFlop/}}}

\begin{document}
\maketitle

\begin{abstract}
Arabic poetry is one of the richest and most culturally rooted forms of expression in the Arabic language, known for its layered meanings, stylistic diversity, and deep historical continuity. Although large language models (LLMs) have demonstrated strong performance across languages and tasks, their ability to understand Arabic poetry remains largely unexplored. In this work, we introduce \emph{Fann or Flop}, the first benchmark designed to assess the comprehension of Arabic poetry by LLMs in 12 historical eras, covering 14 core poetic genres and a variety of metrical forms, from classical structures to contemporary free verse. The benchmark comprises a curated corpus of poems with explanations that assess semantic understanding, metaphor interpretation, prosodic awareness, and cultural context. We argue that poetic comprehension offers a strong indicator for testing how good the LLM understands classical Arabic through Arabic poetry. Unlike surface-level tasks, this domain demands deeper interpretive reasoning and cultural sensitivity. Our evaluation of state-of-the-art LLMs shows that most models struggle with poetic understanding despite strong results on standard Arabic benchmarks. We release \emph{Fann or Flop} \footnote{\href{https://huggingface.co/datasets/omkarthawakar/FannOrFlop}{https://huggingface.co/datasets/omkarthawakar/FannOrFlop}} along with the evaluation suite \footnote{\href{https://github.com/mbzuai-oryx/FannOrFlop}{https://github.com/mbzuai-oryx/FannOrFlop}} as an open-source resource to enable rigorous evaluation and advancement for Arabic language models.
\def\thefootnote{$\dagger$}\footnotetext{Equal contribution.}
\end{abstract}

\begin{figure}[t]
\centering
\includegraphics[width=0.9\linewidth,height=7cm]{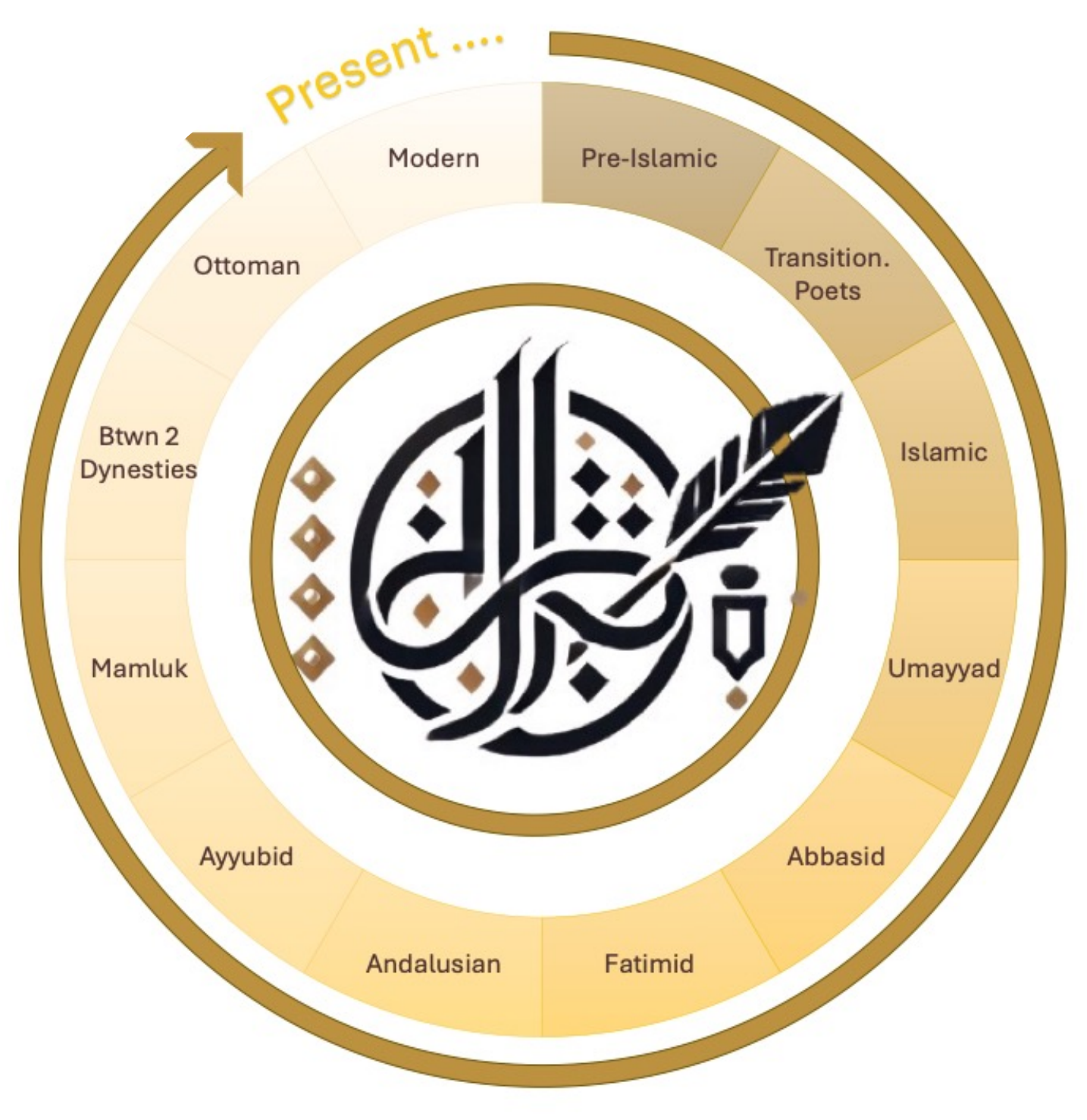}

\caption{
\textbf{Chronological Wheel of Arabic Poetic Eras.} This circular taxonomy visualizes the evolution of Arabic poetry across 12 major historical eras, from the Pre-Islamic and Transitional periods through the Abbasid, Andalusian, and Mamluk dynasties, up to the Modern era. The layout reflects both temporal flow and the rich cultural shifts that shaped poetic expression. Detailed taxonomy by genre, meter, and notable poets presented in Table~\ref{tab:taxonomy}.
}
\label{fig:taxonomy}
\end{figure}

\begin{table*}[ht]
\centering
\setlength{\tabcolsep}{0.5em}
\renewcommand{\arraystretch}{1.25}
\resizebox{\textwidth}{!}{
\begin{tabular}{l|cccccccccccc}
\toprule
\rowcolor{gray!10}
\textbf{Feature} & 
\makecell[c]{\textbf{AQMAR}} & 
\makecell[c]{\textbf{Tafsir}} & 
\makecell[c]{\textbf{Ashaar}} & 
\makecell[c]{\textbf{Ara}\\\textbf{Bench}} & 
\makecell[c]{\textbf{Arabic}\\\textbf{SQuAD}} & 
\makecell[c]{\textbf{ARCD}} & 
\makecell[c]{\textbf{AraBERT}\\\textbf{Collection}} & 
\makecell[c]{\textbf{CAMeL}\\\textbf{Corpus}} & 
\makecell[c]{\textbf{Tash}\\\textbf{keela}} & 
\makecell[c]{\textbf{PADIC}} & 
\makecell[c]{\textbf{MADAR}} & 
\makecell[c]{\textbf{Fann or}\\\textbf{Flop}} \\
\midrule
Dialectal Variety               & \xmark & \xmark & \xmark & \cmark & \xmark & \xmark & \cmark & \xmark & \xmark & \cmark & \cmark & \cmark \\
Poetic Device Annotation        & \xmark & \xmark & \xmark & \xmark & \xmark & \xmark & \xmark & \xmark & \xmark & \xmark & \xmark & \cmark \\
Verse/Sentence-Level Annotation & \xmark & \cmark & \cmark & \xmark & \xmark & \xmark & \xmark & \xmark & \xmark & \xmark & \xmark & \cmark \\
Temporal/Historical Context     & \xmark & \cmark & \cmark & \xmark & \xmark & \xmark & \xmark & \xmark & \xmark & \xmark & \xmark & \cmark \\
QA-Style Task Format            & \xmark & \xmark & \xmark & \xmark & \cmark & \cmark & \cmark & \xmark & \xmark & \xmark & \xmark & \cmark \\
Open-Source                     & \cmark & \cmark & \cmark & \cmark & \cmark & \cmark & \cmark & \cmark & \cmark & \cmark & \cmark & \cmark \\
\bottomrule
\end{tabular}}
\vspace{-0.5em}
\caption{
\textbf{Comparison of key Arabic NLP datasets.} Existing Arabic NLP resources typically address isolated features such as dialectal coverage, QA formats, or classical text processing. In contrast, \textbf{Fann or Flop} uniquely integrates multiple underrepresented dimensions (i.e. dialectal diversity, poetic device annotation, verse-level granularity, temporal grounding, and a QA-style evaluation format) positioning it as the first comprehensive benchmark for Arabic poetry understanding. AQMAR~\cite{mohit2012recall}, Tafsir~\cite{ahmed-etal-2022-tafsir}, Ashaar~\cite{alyafeai2023ashaar}, AraBench~\cite{sajjad2020arabench}, Arabic-SQuAD~\cite{mozannar2019neural}, ARCD~\cite{mozannar2019neural}, AraBERT Collection~\cite{antoun2020arabert}, CAMeL Corpus~\cite{abdul2020arbert,khalifa-etal-2018-morphologically}, Tashkeela~\cite{zerrouki2017tashkeela}, PADIC~\cite{meftouh2015machine}, MADAR~\cite{bouamor2018madar}.
}
\label{tab:dataset_comparison_all}
\end{table*}

\section{Introduction}
Arabic is among the world’s most lexically rich languages, with a vocabulary exceeding 12.3 million words—far surpassing that of most modern languages~\cite{AlSuyuti,ititranslatesArabicRichest}. A single word can convey multiple meanings, varied pronunciations, and diverse interpretations, reflecting the language’s profound semantic complexity. Despite its official status in 27 countries—ranking third in global geopolitical presence~\cite{wikipedia}—only a fraction of this lexicon remains in common use today.

To unify communication across its many dialects, Modern Standard Arabic (MSA) emerged in the late 19th and early 20th centuries as a formal register~\cite{jarrousseModernStandard}. Today, it is the primary language of education, media, and governance in the Arab world. Although linguists distinguish Classical Arabic (CA) from MSA, native speakers generally view them as a unified formal variety~\cite{wikipediaVarietiesArabic}. Nevertheless, even the most comprehensive Arabic dictionaries—such as Lisan al-Arab~\cite{IbnManzur}, Taj al-Lugha~\cite{AlJawhari}, and al-Mu‘jam al-Mu‘asir~\cite{alsharekh,ksaa}—cover only a small portion of the historical corpus, revealing the inherent challenges of Arabic lexicography and sociolinguistic narrowing of usage.
Within this broader linguistic context, Arabic poetry has served as a repository of cultural and intellectual expression from the older era to the modern time. Poetic forms such as long odes (qasida), lyrical love poems (ghazal), elegies (ritha'), strophic songs (muwashsha), and vernacular verse (zajal) are marked by distinct metrical, rhetorical, and performative characteristics. 
While contemporary poets explore free verse and modernist motifs, classical forms continue to exert a strong literary and cultural influence.

Recent advances in LLM, such as GPT~\cite{chen2025sharegpt4v}, LLaMA~\cite{touvron2023llama}, AceGPT~\cite{huang2023acegpt}, Jais~\cite{sengupta2023jais}, and Falcon~\cite{malartic2024falcon2}, have demonstrated impressive multilingual capabilities, including Arabic. However, most Arabic natural language processing (NLP) benchmarks focus on tasks such as sentiment analysis, question answering, or recognition of named entities~\cite{antoun2020arabert,abdul-mageed-etal-2021-dialex,obeid2020camel}, typically in MSA or dialectal prose. These benchmarks often miss the linguistic depth and cultural nuances that are inherent in Arabic poetry.
As LLMs are increasingly evaluated for their ability to handle complex linguistic phenomena, such as metaphor, figurative language, and stylistic nuance, their limitations become evident~\cite{liu2022testing,bisk2020piqa}. The FLUTE benchmark \cite{chakrabarty2022flute} and the FigLang 2024 workshop \cite{googleFigLang2024} have reaffirmed that non-literal language understanding remains a significant challenge. This challenge is particularly acute in Arabic, where poetry is densely layered with intertextuality and cultural symbolism. Arabic poetry thus provides a uniquely demanding testbed for assessing deep linguistic in language models.

\begin{figure*}[t!]
\centering  
\includegraphics[width=\textwidth]{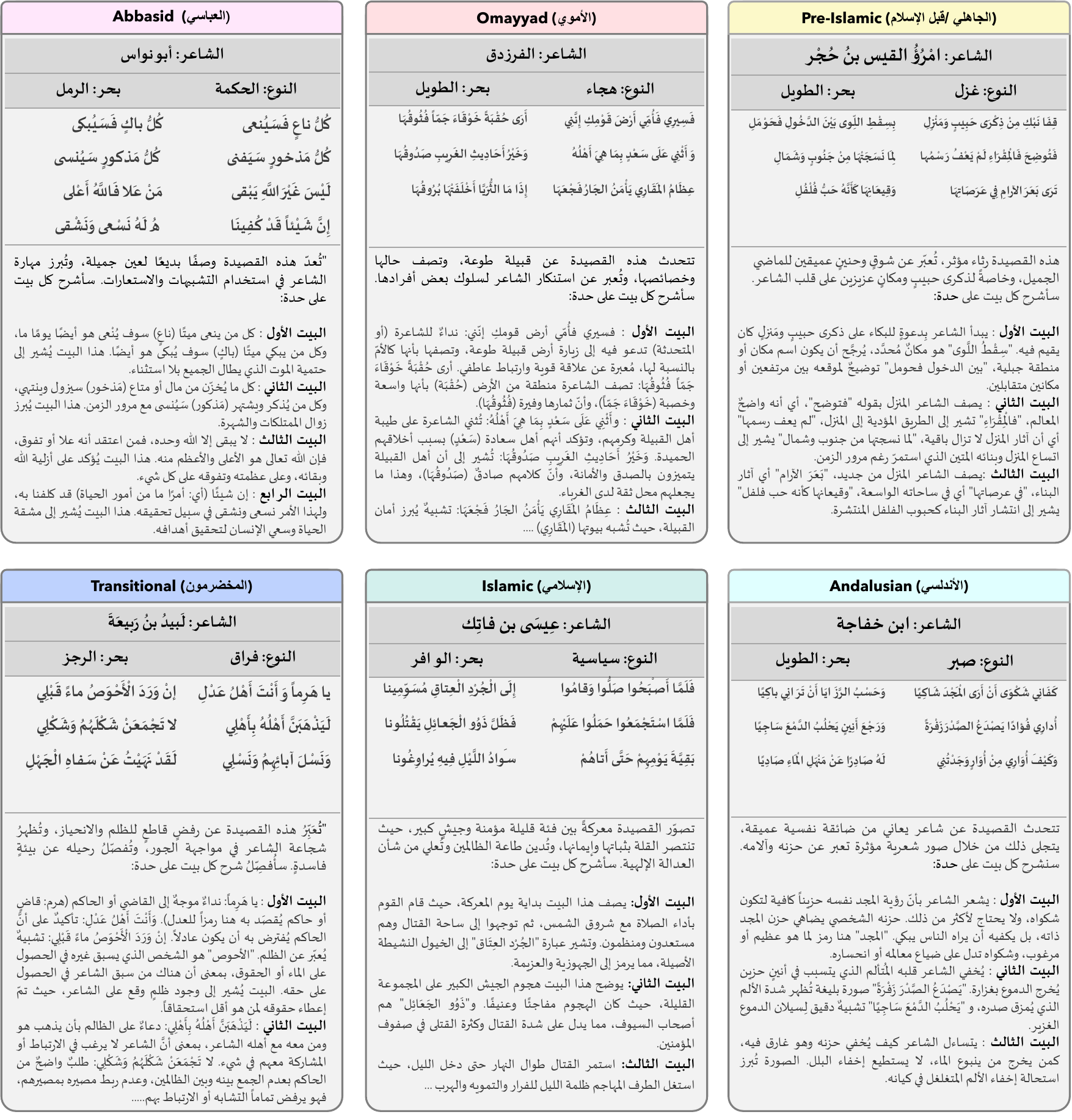}
\vspace{-1.0em}
\caption{
\textbf{Representative Poetic Samples Across Arabic Literary Eras.} This figure presents curated excerpts from Arabic poems spanning key historical eras, illustrating the evolution of language, themes, and stylistic expression. The Pre-Islamic sample reflects tribal valor and rhetorical precision; the Umayyad excerpt captures satire and social commentary; the Abbasid example highlights philosophical reflection and refined metaphorical use; the Transitional era showcases a poetic voice confronting injustice and advocating moral clarity; and the Andalusian selection reveals emotional openness and psychological depth through lyrical expression. Together, these samples provide insight into how Arabic poetry has adapted to diverse historical, cultural, and ideological contexts. Refer to Appendix~\ref{app_en_trans}, Figure~\ref{fig:era_sample_en} for the GPT-4o-generated English translations of the Arabic poetic samples.
}
  \label{fig:dataset_sample}
  \vspace{-0.5em}
\end{figure*}

To address this gap, we introduce \textit{Fann or Flop}, the first benchmark dedicated to evaluating LLMs' understanding of Arabic poetry. Our benchmark comprises 6,984 poem-explanation pairs curated from 12 distinct historical poetic eras (see Figure~\ref{fig:taxonomy}), which can be broadly seen as spanning three major historical periods: pre-Islamic, classical, and contemporary. It covers 14 poetic genres and includes a range of metrical forms, as detailed in Table~\ref{fig:taxonomy}. Each sample is manually verified by native Arabic speakers with domain knowledge to ensure linguistic authenticity and interpretive accuracy. This rich and diverse collection makes Fann or Flop a reliable benchmark for evaluating deep cultural and literary reasoning in Arabic NLP. Figure~\ref{fig:dataset_sample} represents the examples from our proposed Fann or Flop dataset, showcasing the diversity of eras, genres, and poetic styles covered.

\begin{table*}[t!]
\centering
\resizebox{\textwidth}{!}{
\begin{tabular}{p{3cm}|p{2.5cm}|p{3cm}|p{3.5cm}|p{4.5cm}}
\toprule
\rowcolor{gray!10} \textbf{Era} & \textbf{Approx. Years} & \textbf{Genres (Theme)} & \textbf{Meter} & \textbf{Notable Poets} \\
\midrule
\textbf{Pre-Islamic (Jahiliyyah)} & Until 610 CE & Satire, Separation, Wisdom & At-Tawil, Al-Kamel, Al-Basit & Imru al-Qays, Antarah ibn Shaddad, Zuhayr ibn Abi Sulma \\
\midrule
\textbf{Transitional Poets (Mukhadramun)} & Late 6th -- Early 7th c. & Praise, Apology, Religious & Ar-Rojz, Ar-Ramel & Hassan ibn Thabit, Labid ibn Rabi’a, Al-Khansa\\
\midrule
\textbf{Islamic} & 610--661 CE & Religious, Wisdom, Patience & Al-Madid, Al-Kamel & Abu Sallama Al-Arhabi, Onayf Ibn Kitra \\
\midrule
\textbf{Umayyad} & 661--750 CE & Love, Satire, Political & At-Tawil, Al-Wafer, As-Sari' & Jarir, al-Farazdaq, al-Akhtal\\
\midrule
\textbf{Abbasid} & 750--1258 CE & Praise, Elegy, Wisdom & Al-Basit, Kamel, Al-Monsareh, Al-Moktadab & Abu Nuwas, al-Mutanabbi, al-Buhturi, Abu Tammam \\
\midrule
\textbf{Fatimid} & 909--1171 CE & Religious, Praise, Sadness & Ar-Rojz, Al-Mutakareb & Ibn Hayus, Abu al-Ala al-Ma'arri \\
\midrule
\textbf{Andalusian} & 756--1492 CE & Love, Longing, Wisdom & Mowachah, Al-Mowaliya, Al-Mohtath & Ibn Sahl Al-Andalusi, Ibn Zaydun, Ibn Khafaja \\
\midrule
\textbf{Ayyubid} & 1171--1250 CE & Religious, Praise, Elegy & Al-Kamel, Al-Khafif & Ibn al-Farid, Mohyiddine Bin Arabi \\
\midrule
\textbf{Mamluk} & 1250--1517 CE & Wisdom, Praise, Religious & Al-Wafer, Ar-Rojz & Bahaa'eddine Zuhair, Safiyueddine Alhilli\\
\midrule
\textbf{Between the Two Dynasties} & 1258--1517 CE & Religious, Wisdom, Reproach & Al-Mutadarek, Ar-Ramel & Bashar bn Burd \\
\midrule
\textbf{Ottoman} & 1517--1800 CE & Religious, Love, General & Al-Kamel, Al-khafif & Bnt Al-Shahna, Ibn Razka \\
\midrule
\textbf{Modern} & 19th c. -- Present & Nationalism, Love, Social Justice & Free Meter & Ahmad Shawqi, Hafeth Ibrahim \\
\bottomrule
\end{tabular}
}
\vspace{-0.5em}
\caption{
\textbf{Taxonomy of Arabic Poetic Eras with Genre and Meter Coverage.} 
This table provides a structured overview of 12 major eras in Arabic poetic history, detailing their approximate chronological spans, the most prominent poetic themes (genres) representative of each era, the dominant metrical patterns (Arabic \textit{buḥūr}) used in poetic composition, and notable poets who exemplify the literary character of their time. The genre column highlights recurring thematic concerns such as satire, elegy, love, nationalism, and religious devotion, while the meter column showcases the classical metrical forms like \textit{At-Tawil}, \textit{Al-Kamel}, and \textit{Ar-Rojz}, along with innovations such as free verse in the modern period. This taxonomy reflects the dynamic interplay between form, content, and historical context in shaping Arabic poetic expression.}
\label{tab:taxonomy}
\end{table*}

Our goal is to provide a diagnostic on how well your language model understands and interprets genuine Classical Arabic. Unlike general text, poetry requires sensitivity to rhetorical devices, metrical patterns, and sociohistorical context, making it a rigorous and culturally grounded indicator of language proficiency. Our benchmark serves as a clear indicator of whether a model has truly been exposed to and internalized high-quality Arabic content, offering a focused lens into its cultural and linguistic depth.
We evaluate a range of open-source and commercial LLMs using Fann or Flop and find that, despite strong performance on conventional Arabic tasks, most models struggle with the interpretive depth required by poetry. These findings highlight the need for culturally informed benchmarks that better reflect the depth and diversity of Arabic. We release Fann or Flop as an open-source resource to support the development and evaluation of Arabic-capable language models.

\vspace{1.0em}

\section{The Fann or Flop Dataset}
\subsection{Dataset Taxonomy}
\label{sec:data_tax}
To capture the linguistic, historical, and thematic richness of Arabic poetry, we construct an expert-verified taxonomy that organizes poems across both form and era. As illustrated in Figure~\ref{fig:taxonomy} and detailed in Table~\ref{tab:taxonomy}, the taxonomy traces 12 distinct poetic eras, from the pre-Islamic period to modern times, encompassing 14 genres that capture the dominant styles, concerns, and historical contexts of each era. It illustrates how poetic expression evolved over the centuries.

This structured framework was carefully reviewed and validated by scholars specializing in Arabic language and literature to ensure both linguistic accuracy and contextual relevance. Their expertise helped align the taxonomy with established literary traditions while accommodating the nuances of classical and modern poetic forms. Beyond its utility for literary and philological analysis, the taxonomy serves as a robust foundation for computational modeling. It enables more precise automatic genre classification and facilitates temporal contextualization across different eras of Arabic poetry, thereby supporting culturally informed and interpretable Arabic NLP research.

\begin{figure*}[t!]
\centering
  \includegraphics[width=\textwidth,height=5.5cm]{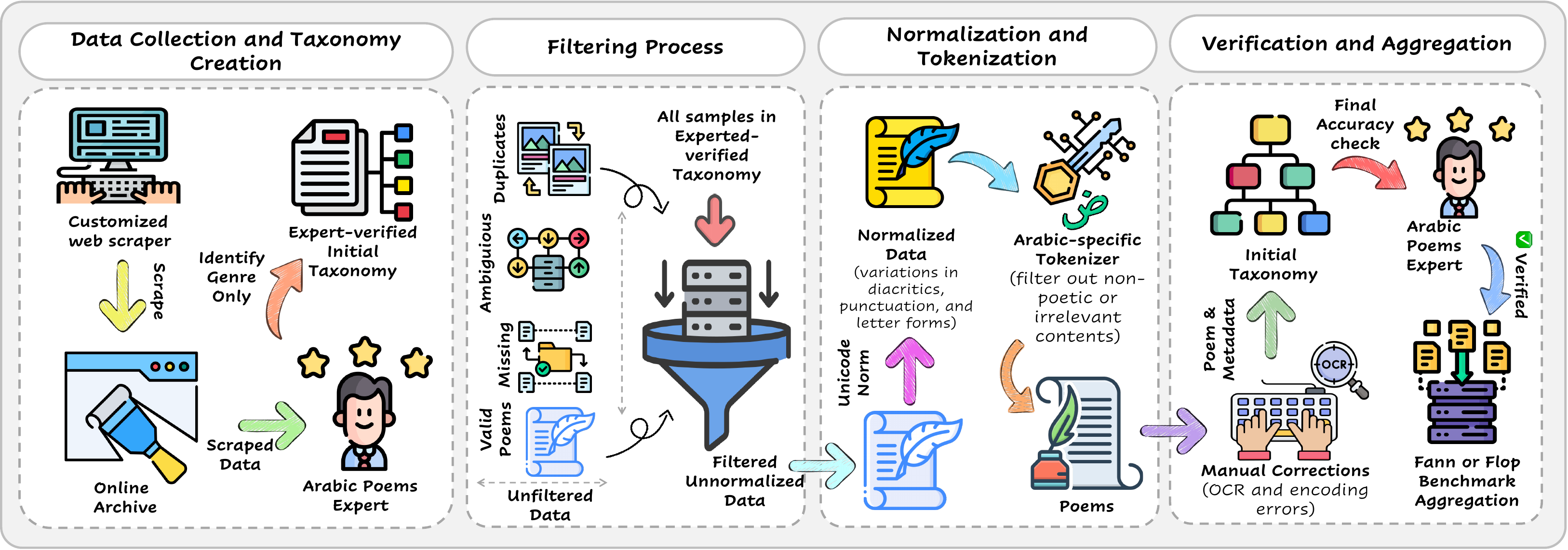}
\vspace{-1.0em}
\caption{
\textbf{Fann or Flop Pipeline.} Fann or Flop is built out of the multi-stage pipeline. It begins with scraping Arabic poems
from a trusted online archive using a custom web scraper. Extracted poems are matched to an initial expert-verified taxonomy
and filtered to remove duplicates, ambiguous metadata, and invalid entries. The filtered texts then undergo normalization
(e.g., unifying diacritics, punctuation, and letter forms) and Arabic-specific tokenization, with non-poetic or irrelevant content
excluded. Manual corrections are applied to fix OCR and encoding errors. In the final stage, linguistic experts verify
each sample to ensure proper alignment with genre and era labels.
}
\label{fig:data_pipeline}
\vspace{-1em}
\end{figure*}

\subsection{Data Collection}
We curated Arabic poems from a well-established digital archive\footnote{\url{https://arabic-poetry.net}}, which hosts a broad spectrum of poets, genres, and historical periods. A custom web scraper was developed to extract the poem texts along with associated metadata, including the poet's name, historical era, genre, meter, and poem's title. The resulting dataset extends across 12 distinct eras, from the pre-Islamic period to the modern era, and reflects a diverse range of poetic forms and styles. All entries were curated following our expert-verified taxonomy (see Table~\ref{tab:taxonomy}), ensuring consistency across genre and era classifications. This structured approach preserves both the linguistic richness and historical specificity of Arabic poetry, providing a valuable resource for research in both Arabic NLP and digital literary studies. By aligning each sample with a well-defined literary context, the dataset enables more accurate model evaluation and supports culturally grounded language understanding.

\subsection{Data Filtering and Verification}

To ensure data quality, consistency, and alignment with the expert-defined taxonomy, we applied a multi-step filtering and verification pipeline, illustrated in Figure~\ref{fig:data_pipeline}. The process consisted of the following stages:

\begin{itemize}
    \item [-]\textbf{Duplicate and Metadata Filtering:} Starting with a collection of over 10,000 Arabic poems, we removed duplicate entries and discarded those with missing or ambiguous metadata, such as unknown poets or unspecified historical eras, resulting in a curated dataset of 6,984 high-quality poems.
    \item [-]\textbf{Unicode Normalization:} All poems were standardized using Unicode normalization to address orthographic inconsistencies common in Arabic, including variations in diacritics, punctuation, and letter forms (e.g., alternate representations of \emph{alif} and \emph{ta marbuta}).
    \item [-]\textbf{Text Tokenization and Content Filtering:} We applied an Arabic-specific tokenizer to segment the text accurately. Non-poetic or irrelevant content, such as editorial comments, footnotes, and prose fragments, was automatically excluded.
    \item [-]\textbf{Manual Correction of Encoding Errors:} A sample subset of poems was manually reviewed to correct common OCR and encoding issues that were not resolved through automated preprocessing.  
    \item [-] \textbf{Expert Validation of Labels:} All genre and era annotations were reviewed by Arabic language and literature experts. This validation step ensured that each poem was accurately categorized in accordance with the taxonomy introduced in Section~\ref{sec:data_tax}.
\end{itemize}

\begin{table*}[t!]
\centering
\small
\setlength{\tabcolsep}{3pt}
\renewcommand{\arraystretch}{1.15}
    \resizebox{\textwidth}{!}{%
    \begin{tabular}{llccccccc}
    \toprule
    \rowcolor{gray!10} & \textbf{Model} & \textbf{BLEU} & \textbf{chrF(++)} & \textbf{BERTScore} &\textbf{Textual}& \textbf{Faithfulness/} & \textbf{Fluency/} & \textbf{Interpretive-} \\
    \rowcolor{gray!10} & &  &  &  &\textbf{Entailment}& \textbf{Consistency} & \textbf{Grammaticality} & \textbf{Depth} \\
    \midrule
    \multirow{6.2}{*}{
    \begin{tabular}[c]{@{}c@{}}
    \rotatebox{90}{\small{Closed}}\end{tabular}}
    & GPT-4o-2024-08-06~\cite{openai2024gpt4ocard} & 0.0395 & \textbf{0.2882} & \textbf{0.6410} & 0.6775 & 3.92 ($\pm$ 0.99) & 4.96 ($\pm$0.20) & \textbf{7.52} \\
    & GPT-4o-mini-2024-07-18~\cite{gpt4omini} & 0.0395 & 0.2542 & 0.6124 & 0.4383 & 2.91 ($\pm$0.75) & 4.28 ($\pm$0.57) & 7.50 \\
    & Gemini-2.5-Flash~\cite{gemini2.5flash} & 0.0153 & 0.2618 & 0.6319 & \textbf{0.7475} & \textbf{4.25} ($\pm$1.00) & \textbf{4.98} ($\pm$0.16) & 7.22 \\
    & Gemini-2.0-Flash~\cite{gemini2.0flash} & 0.0395 & 0.2618 & 0.6393 & 0.7154 & 3.99 ($\pm$1.04) & 4.95 ($\pm$0.22) & 6.50 \\
    & Gemini-1.5-Pro~\cite{gemini1.5} & 0.0395 & 0.2618 & 0.6333 & 0.6180 & 3.59 ($\pm$1.00) & 4.80 ($\pm$0.41) & 5.38 \\
    & Fanar-Star~\cite{fanar} & 0.0138 & 0.1538 & 0.5677 & 0.6468 & 2.16 ($\pm$0.92) & 3.40 ($\pm$0.76) & 2.88 \\
    \midrule
    \multirow{9.2}{*}{
    \begin{tabular}[c]{@{}c@{}}
    \rotatebox{90}{\small{Open}}\end{tabular}}
    & Deepseek-V3~\cite{liu2024deepseek} & 0.0395 & 0.2771 & 0.6335 & 0.5117 & 3.36 ($\pm$0.91) & \textbf{4.98} ($\pm$0.16) & 4.75 \\
    & Deepseek-R1~\cite{guo2025deepseek} & 0.0395 & 0.2771 & 0.6335 & 0.5117 & 3.38 ($\pm$0.92) & \textbf{4.98} ($\pm$0.16) & 4.25 \\
    & Llama-3.3-70B~\cite{llama3.3} & 0.0153 & 0.2618 & 0.6393 & 0.5364 & 2.51 ($\pm$0.90) & 3.37 ($\pm$0.73) & 7.20 \\
    & Qwen-3~\cite{qwen3} & 0.0296 & 0.2837 & 0.6158 & 0.6468 & 3.98 ($\pm$0.90) & 4.73 ($\pm$0.45) & 6.50 \\
    & Aya-Expanse~\cite{dang2024aya} & 0.0329 & 0.2771 & 0.6328 & 0.6468 & 3.76 ($\pm$0.90) & 4.68 ($\pm$0.47) & 5.88 \\
    & Jais~\cite{sengupta2023jais} & 0.0312 & 0.2698 & 0.6245 & 0.6023 & 3.21 ($\pm$0.88) & 4.35 ($\pm$0.52) & 5.35 \\
    & ALLaM-7B~\cite{allam} & 0.0119 & 0.0463 & 0.5375 & 0.5997 & 1.32 ($\pm$0.62) & 2.11 ($\pm$0.89) & 3.12 \\
    & AceGPT-v2-70B-Chat~\cite{huang2023acegpt} & \textbf{0.0402} & 0.0412 & 0.5759 & 0.6061 & 2.52 ($\pm$0.91) & 3.46 $\pm$0.95) & 4.12 \\
    \bottomrule
    \end{tabular}
    }
    \vspace{-0.5em}
\caption{
\textbf{Comparison of closed and open-source models on the Arabic poem understanding task using both automatic and human evaluations.} BLEU, chrF(++), and BERTScore capture lexical and semantic similarity with reference explanations, while textual entailment assesses factual alignment. Human evaluation includes interpretive depth, while faithfulness and fluency are automatically judged using GPT-4o as a reference grader. Closed models like GPT-4o and Gemini-2.5-Flash achieve strong overall performance, while open models such as Deepseek-V3 and Aya-Expanse show promising consistency and interpretability. This benchmark highlights the potential of open models and the need for deeper cultural reasoning in Arabic poetic understanding.
} 
\label{tab:model_performance}
\end{table*}

 \begin{table*}[t!]
\centering
\small
\setlength{\tabcolsep}{6pt}
\renewcommand{\arraystretch}{1.15}
\resizebox{\textwidth}{!}{%
\begin{tabular}{llcccccc}
\toprule
\rowcolor{gray!10} & \textbf{Model} & \textbf{Pre-Islamic} & \textbf{Transitional} & \textbf{Early Islamic} & \textbf{Umayyad} & \textbf{Abbasid} & \textbf{Fatimid} \\
\midrule
\multirow{6.2}{*}{\rotatebox{90}{\small{Closed}}} 
 & GPT-4o-2024-08-06~\cite{openai2024gpt4ocard} & 0.6285 & 0.6304 & 0.6341 & 0.6285 & 0.6421 & 0.6398 \\
    & GPT-4o-mini-2024-07-18~\cite{gpt4omini} & 0.5980 & 0.6060 & 0.6134 & 0.5998 & 0.6125 & 0.6127  \\
    & Gemini-2.5-Flash~\cite{gemini2.5flash} & 0.6245 & 0.6264 & 0.6286 & 0.6253 & 0.6326 & 0.6282 \\
    & Gemini-2.0-Flash~\cite{gemini2.0flash} & 0.6290 & 0.6303 & 0.6326 & 0.6312 & 0.6404 & 0.6373 \\
    & Gemini-1.5-Pro~\cite{gemini1.5} & 0.6255 & 0.6293 & 0.6223 & 0.6278 & 0.6338 & 0.6307 \\
    & Fanar-Star~\cite{fanar} & 0.5694 & 0.5749 & 0.5695 & 0.5696 & 0.5720 & 0.5666 \\
\midrule
\multirow{9.2}{*}{\rotatebox{90}{\small{Open}}}
    & Deepseek-V3~\cite{liu2024deepseek} & 0.6225 & 0.6303 & 0.6311 & 0.6263 & 0.6313 & 0.6330 \\
    & Deepseek-R1~\cite{guo2025deepseek} & 0.6271 & 0.6296 & 0.6321 & 0.6247 & 0.6324 & 0.6359 \\
    & Llama-3.3-70B~\cite{llama3.3} & 0.5705 & 0.5703 & 0.5701 & 0.5668 & 0.5831 & 0.5719 \\
    & Qwen-3~\cite{qwen3} & 0.6111 & 0.6152 & 0.6129 & 0.6136 & 0.6164 & 0.6145 \\
    & Aya-Expanse~\cite{dang2024aya} & 0.6214 & 0.6232 & 0.6220 & 0.6232 & 0.6343 & 0.6294 \\
    & Jais~\cite{sengupta2023jais} & 0.6172 & 0.6218 & 0.6241 & 0.6183 & 0.6285 & 0.6239 \\
    & ALLaM-7B~\cite{allam} & 0.5786 & 0.5826 & 0.5917 & 0.5790 & 0.5862 & 0.5799 \\
    & AceGPT-v2-70B-Chat~\cite{huang2023acegpt} & 0.6194 & 0.6246 & 0.6329 & 0.6213 & 0.6261 & 0.6225   \\
\bottomrule
\end{tabular}}

\vspace{1em}

\resizebox{\textwidth}{!}{%
\begin{tabular}{llcccccc}
\toprule
\rowcolor{gray!10} & \textbf{Model} & \textbf{Andalusian} & \textbf{Ayyubid} & \textbf{Mamluk} & \textbf{Between Dynasties} & \textbf{Ottoman} & \textbf{Modern } \\
\midrule
\multirow{4.2}{*}{\rotatebox{90}{\small{Closed}}} 
& GPT-4o-2024-08-06~\cite{openai2024gpt4ocard} & 0.6386 & 0.6440 & 0.6563 & 0.6440 & 0.6510 & 0.6487 \\
    & GPT-4o-mini-2024-07-18~\cite{gpt4omini} & 0.6151 & 0.6167 & 0.6273 & 0.6176 & 0.6202 & 0.6140  \\
    & Gemini-2.5-Flash~\cite{gemini2.5flash} & 0.6297 & 0.6340 & 0.6421 & 0.6336 & 0.6415  & 0.6341 \\
    & Gemini-2.0-Flash~\cite{gemini2.0flash} & 0.6346 & 0.6409 & 0.6533 & 0.6414 & 0.6504 & 0.6441 \\
    & Gemini-1.5-Pro~\cite{gemini1.5} & 0.6313 & 0.6349 & 0.6409 & 0.6355 & 0.6443 & 0.6387 \\
    & Fanar-Star~\cite{fanar} & 0.5746 & 0.5684 & 0.5569 & 0.5831 & 0.5586 & 0.5392 \\
\midrule
\multirow{9.2}{*}{\rotatebox{90}{\small{Open}}}
    & Deepseek-V3~\cite{liu2024deepseek} & 0.6337 & 0.6404 & 0.6482 & 0.6393 & 0.6404 & 0.6368 \\
    & Deepseek-R1~\cite{guo2025deepseek} & 0.6353 & 0.6404 & 0.6509 & 0.6408 & 0.6423 & 0.6373 \\
    & Llama-3.3-70B~\cite{llama3.3} & 0.5791 & 0.5755 & 0.5935 & 0.5854 & 0.5797 & 0.5794 \\
    & Qwen-3~\cite{qwen3} & 0.6153 & 0.6163 & 0.6189 & 0.6160 & 0.6242 & 0.6149 \\
    & Aya-Expanse~\cite{dang2024aya} & 0.6289 & 0.6366 & 0.6475 & 0.6367 & 0.6393 & 0.6398 \\
    & Jais-30B-v3~\cite{sengupta2023jais} & 0.6279 & 0.6321 & 0.6413 & 0.6307 & 0.6348 & 0.6316 \\
    & ALLaM-7B~\cite{allam} & 0.5876 & 0.5925 & 0.6004 & 0.5884 & 0.5933 & 0.5864 \\
    & AceGPT-v2-70B-Chat~\cite{huang2023acegpt} & 0.6168 & 0.6280 & 0.6466 & 0.6212 & 0.6205 & 0.6265  \\
\bottomrule
\end{tabular}}

\vspace{-0.5em}
\caption{
\textbf{Era-wise Evaluation using BERTScore.} Model-wise performance breakdown using BERTScore evaluation across different Arabic poetic eras, evaluating understanding and generation quality within historical and stylistic contexts. The eras span from Pre-Islamic to Modern periods, offering a fine-grained analysis of model capabilities across evolving linguistic and cultural expressions. This table highlights gaps in temporal generalization and cultural grounding, motivating the need for era-aware training and evaluation in Arabic literary modeling.
}
\label{tab:performance_era}
\end{table*}

\section{Fann or Flop Benchmark Evaluation}

\noindent\textbf{Evaluation Metric:}
To evaluate the quality of LLM-generated explanations for Arabic poetry, we employ a multi-tiered evaluation framework combining automatic metrics, semantic and entailment-based modeling, LLM-as-Judge scoring, and human expert annotation. This design enables us to capture both surface-level fidelity and the deeper interpretive demands of poetic understanding.

For automatic evaluation, we compute BLEU~\cite{papineni2002bleu} and chrF(++)~\cite{popovic-2017-chrf} scores to quantify semantic and character-level overlap between model outputs and actual poem explanation references. While useful for consistency checks, these metrics are limited in capturing the nuanced variation allowed in literary interpretation.

To assess semantic alignment, we employ BERTScore~\cite{zhang2019bertscore}, leveraging Arabic-pretrained transformers such as AraBERT~\cite{antoun2020arabert} to quantify the semantic similarity between model-generated explanations and human-authored references. In addition, we incorporate Textual Entailment (TE) analysis using mDeBERTaV3~\cite{he2021debertav3}, a multilingual model fine-tuned for Natural Language Inference (NLI). This enables us to evaluate whether the reference explanation logically entails the generated output. Unlike surface-level similarity metrics, this approach provides a deeper measure of semantic consistency, capturing whether the generated interpretation remains faithful to the intended meaning of the expert-authored reference, even when expressed using different lexical or syntactic forms.

For LLM-as-Judge evaluation, we assess two key aspects: faithfulness/consistency and fluency/grammaticality. Faithfulness evaluates whether the explanation accurately reflects the content of the poem without introducing hallucinated or misleading elements. Fluency assesses the grammaticality, clarity, and naturalness of the output in Arabic. Both dimensions are scored by GPT-4o~\cite{openai2024gpt4ocard}, a state-of-the-art LLM with strong Arabic language understanding.

\begin{figure*}[t!]
\centering  
\includegraphics[width=\textwidth]{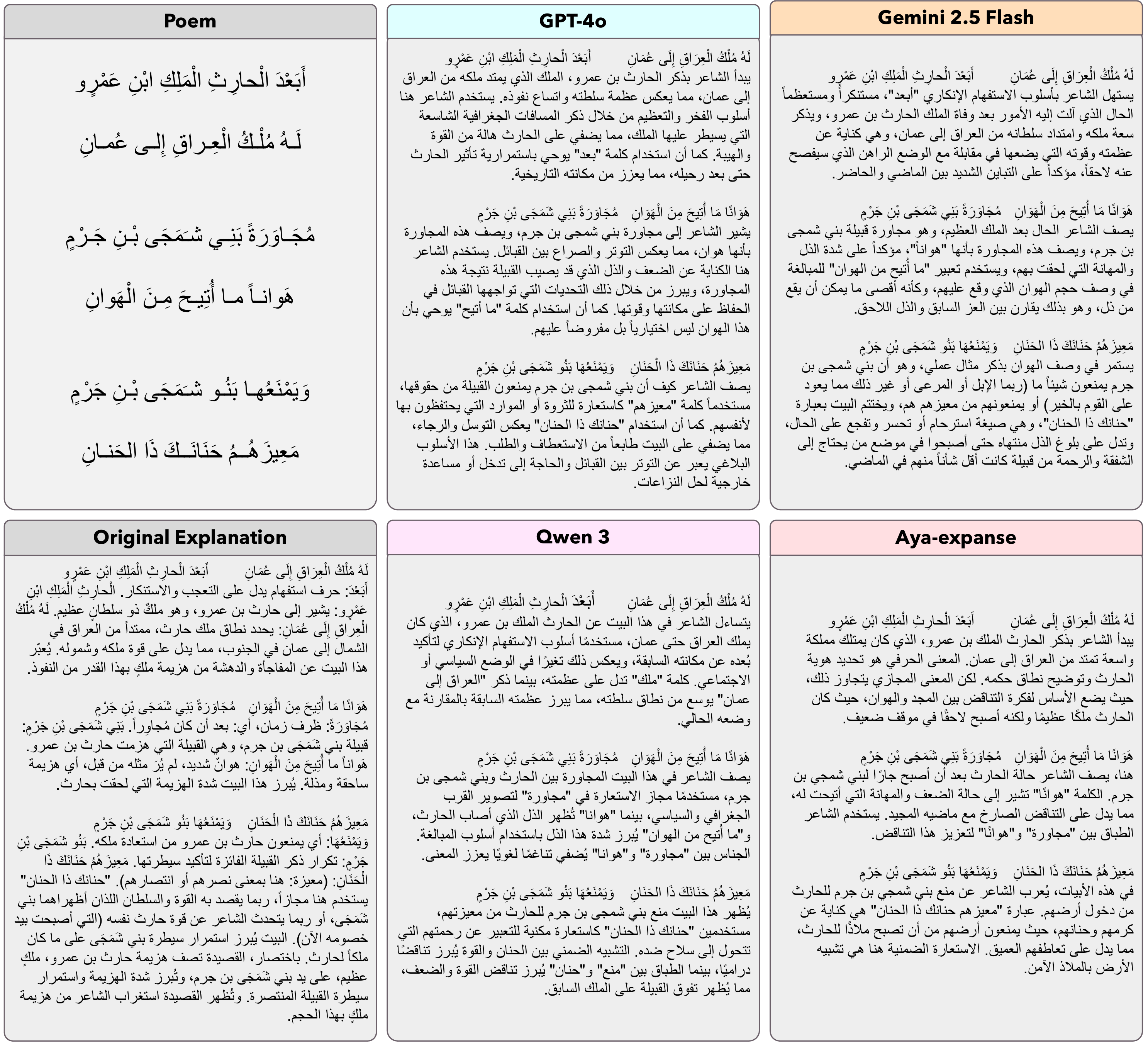}
\vspace{-1.0em}
\caption{
\textbf{Qualitative Comparison of Model-Generated Explanations for a Single Arabic Poem.} This figure presents a representative Arabic poem alongside its original human-written explanation and corresponding verse-by-verse explanations generated by four different language models. The comparison highlights how each model interprets the poem’s rhetorical devices, imagery, and thematic depth relative to the gold explanation. This qualitative analysis illustrates variations in faithfulness, fluency, and literary sensitivity, offering insight into each model’s ability to handle nuanced Arabic poetic language and convey its intended meaning.
}
  \label{fig:qualitative_sample}
  \vspace{-0.5em}
\end{figure*}

To capture the more interpretive and literary dimensions of explanation quality, we conduct human evaluation centered on interpretive depth. Annotators use a rubric-based scale (0–10) grounded in poetic analysis, incorporating the following criteria:
\begin{itemize}
    \item [-] \textbf{Literal Comprehension (0–1):} Does the explanation correctly reflect the surface meaning of the poem?
    \item [-] \textbf{Thematic and Emotional Depth (0–2):} Does it convey underlying themes, sentiment, or tone (e.g., longing, satire, mysticism)?
    \item [-] \textbf{Cultural and Historical Appropriateness (0–2):} Does it demonstrate awareness of cultural, religious, or historical context?
    \item[-] \textbf{Stylistic Sensitivity (0–3):} Does it acknowledge rhetorical and literary features such as metaphor, figurative language, rhythm, or imagery?
    \item [-] \textbf{Expressiveness and Coherence (0–2):} Is the explanation clear, well-articulated, and stylistically appropriate in Arabic?
\end{itemize}
By integrating these complementary evaluation layers, our framework provides a comprehensive and culturally grounded assessment of LLMs’ ability to interpret and explain Arabic poetry.\\

\noindent\textbf{Results and Analysis:}
Table~\ref{tab:model_performance} presents the performance of both closed and open-source models on Arabic poem understanding using a combination of automatic metrics (BLEU, chrF(++), BERTScore, Textual Entailment, faithfulness and fluency) and human evaluation such as interpretive depth analysis. These metrics collectively assess the quality, relevance, and clarity of model-generated explanations when interpreting Arabic poetry.

Overall, closed models such as GPT-4o and Gemini-2.5-Flash achieve consistently strong scores across both automatic and human evaluations. Notably, Gemini-2.5-Flash attains the highest textual entailment score (0.7475), along with high fluency and faithfulness scores, indicating strong alignment with poetic content and natural language clarity. GPT-4o also performs well across all dimensions, with the highest BERTScore and a strong balance of semantic coherence and linguistic quality.
Among open models, Deepseek-V3, Aya-Expanse, and Qwen-3 show competitive performance, especially in fluency and textual entailment. However, models like ALLaM-7B and AceGPT-v2 lag significantly in both lexical and semantic overlap, as well as in human-judged fluency and consistency.

A key insight from this evaluation is that most state-of-the-art models perform well on content expressed in Modern Standard Arabic (MSA) but struggle with the classical forms and linguistic intricacies present in historical and poetic Arabic. Despite high scores in generic semantic metrics, many models fail to capture deeper cultural and metaphorical meanings embedded in traditional Arabic poetry.
Our analysis highlights the importance of domain-specific evaluation for literary and cultural tasks. It also underscores the need for building or fine-tuning models that are more sensitive to classical Arabic forms. The gap between fluency and interpretive depth in some models suggests that future research should focus not just on surface-level correctness but also on deeper reasoning and cultural grounding. Such efforts are essential for advancing Arabic NLP in creative and heritage-preserving applications.

Table~\ref{tab:performance_era} shows era-wise performance of closed and open-source models on Arabic poem understanding using BERTScore, which captures semantic similarity with human explanations. Closed models like GPT-4o and Gemini variants perform consistently well, especially on modern and recent historical eras. In contrast, open models such as Deepseek-V3 and Aya-Expanse perform reasonably on some eras but struggle with older poetic forms like Pre-Islamic and Umayyad due to their complex language and cultural depth. This highlights that while current models are effective on MSA, they face challenges with classical Arabic. A complementary analysis using Textual Entailment is included in the Appendix (refer Table~\ref{tab:performance_era_add}), further supporting these findings.

Additionally, Figure~\ref{fig:qualitative_sample} shows a qualitative comparison of model-generated explanations for a classical Arabic poem. It compares outputs from GPT-4o, Gemini 2.5 Flash, Qwen 3, and Aya-Expanse against a human-written explanation. The figure highlights differences in faithfulness, fluency, and interpretive depth, showing how well each model captures the poem’s meaning, style, and literary richness. This example clearly illustrates the strengths of advanced models like GPT-4o in understanding nuanced poetic language.

\section{Conclusion}

Arabic poetry represents one of the richest and most culturally nuanced forms of expression within the Arabic language, characterized by layered meanings, stylistic diversity, and deep historical roots. In this paper, we introduced Fann or Flop, the first benchmark specifically developed to evaluate the capabilities of LLMs in understanding Arabic poetry across 12 historical eras, spanning from pre-Islamic to contemporary periods, and encompassing a broad spectrum of poetic genres and metrical forms. Our benchmark includes carefully curated diagnostic questions aimed at assessing semantic comprehension, metaphorical interpretation, prosodic awareness, and sensitivity to cultural contexts. Through extensive evaluation, we demonstrated that despite strong performances on standard Arabic language tasks, state-of-the-art LLMs consistently struggle with the interpretative and culturally embedded dimensions of Arabic poetic texts. By releasing Fann or Flop as an open-source resource, we aim to encourage further research, promote rigorous assessment methodologies, and support advancements in linguistically and culturally rich Arabic language modeling.

\section{Limitations and Societal Impact}
While Fann or Flop provides a rigorous framework for evaluating LLMs’ understanding of Arabic poetry, it has several limitations. The benchmark covers only a portion of the broader Arabic poetic tradition, as some poems could not be included due to missing metadata, unclear authorship, or lack of reliable era or genre annotations. Additionally, poetry often invites multiple valid interpretations, which current evaluation metrics may not fully capture, even with expert-curated references. Expanding the dataset to include more diverse annotations, as well as dialectal and regional poetic forms, remains a key area for future work.

On the societal front, this benchmark contributes to the preservation and computational accessibility of Arabic literary heritage by positioning poetry as a meaningful testbed for language understanding. By promoting the development of culturally informed and linguistically grounded models, Fann or Flop encourages more inclusive and context-sensitive NLP. Nonetheless, as with any system trained on culturally rich and potentially sensitive material, there is a risk of misinterpretation or misuse. Ensuring transparency, human oversight, and responsible deployment is essential to safeguard the ethical impact of this work, especially in educational, literary, and public-facing applications.

\bibliography{arxiv}

\clearpage
\appendix

\section{Appendix}
\label{sec:appendix}

This appendix provides supplementary material to support our study of Arabic poetry understanding in language models. It includes four key sections: (1) a brief overview of related work in Arabic NLP, highlighting recent progress in benchmark development and the specific gaps our work addresses; (2) detailed dataset statistics, including token distribution, genre coverage, and temporal representation across poetic eras; (3) additional details on the prompts used for model generation and evaluation; and (4) a selection of qualitative examples from the \emph{Fann or Flop} benchmark that illustrate its richness and the interpretive challenges it presents. Together, these components underscore the linguistic, historical, and cultural depth of our dataset and evaluation framework.

\section{Related Work}

Understanding Arabic poetry computationally intersects with multiple subfields of NLP, including language modeling, data set construction, figurative language interpretation, and the evaluation of cultural knowledge. To contextualize our contribution, we review prior work across two key domains: Arabic NLP benchmarks and poetry understanding in LLMs.

\subsection{Arabic NLP Benchmarks}
Over the past decade, Arabic NLP has advanced considerably with the introduction of large-scale benchmarks such as SOQAL (Arabic-SQuAD and ARCD)~\cite{mozannar2019neural}, AraBench~\cite{sajjad2020arabench}, and the AraBERT Collection~\cite{antoun2020arabert}. These benchmarks cover essential tasks such as sentiment analysis, named entity recognition (NER), and question answering, and typically support both MSA and dialectal varieties. However, they largely overlook CA, which remains underrepresented in the main resources. Consequently, while models trained on these datasets perform well on surface-level tasks, they lack the depth to assess cultural, rhetorical, and literary understanding, especially in classical poetic contexts.

Additional resources such as the CAMeL corpus~\cite{abdul2020arbert,khalifa-etal-2018-morphologically}, Tashkeela~\cite{zerrouki2017tashkeela}, PADIC~\cite{meftouh2015machine}, and MADAR~\cite{bouamor2018madar} have enriched the field through morphologically annotated corpora, diacritized texts, and dialectal content. However, these datasets are primarily designed for structural tasks such as morphological disambiguation or dialect identification, without engaging the semantic or figurative dimensions of the poetic language.

More recently, efforts have extended Arabic NLP to the literary and religious domains. The Tafsir dataset~\cite{ahmed-etal-2022-tafsir} introduces a benchmark derived from \emph{Tafsir al-Tabari}, including NER and topic modeling in CA. AQMAR~\cite{mohit2012recall} targets recall-oriented NER in Arabic Wikipedia, offering annotations across standard and domain-specific entity types. Although both datasets engage with classical Arabic and semantic granularity, they do not address poetry or the interpretive challenges it poses.

Among the most directly relevant efforts is Ashaar~\cite{alyafeai2023ashaar}, the first large-scale Arabic poetry dataset. It includes tasks such as meter classification, era identification, and poet recognition, along with descriptive metadata. Despite its contributions to computational poetics, Ashaar lacks verse-level annotation, rhetorical device modeling, question-answer style interpretation, and historical contextualization, limiting its ability to evaluate deeper poetic reasoning in language models.

\subsection{Poetry Understanding in NLP}
Outside Arabic, poetry and figurative language have emerged as valuable testbeds for assessing the reasoning of LLM~\cite{liu2022testing, bisk2020piqa,olivero2024figurative}. Benchmarks like FLUTE~\cite{chakrabarty2022flute} and the FigLang shared tasks~\cite{googleFigLang2024} reveal persistent challenges in handling metaphor, simile, and symbolic expression. Recent works~\cite{gallipoli2025not, zhao2024understanding} further expose the limitations of LLMs in interpreting literary texts, including complex poetic structures and non-literal meaning. Despite Arabic's longstanding poetic legacy, this evaluation line remains largely unexplored for Arabic, leaving a notable gap in culturally grounded reasoning tasks.

\textbf{Fann or Flop} addresses this gap by combining a chronological taxonomy of Arabic poetry with interpretive question-answering. It spans 12 eras and integrates dialectal variation, rhetorical analysis, historical context, and verse-level annotation. As summarized in Table~\ref{tab:dataset_comparison_all}, no existing benchmark offers this breadth of poetic features, positioning Fann or Flop as the first comprehensive diagnostic tool for evaluating Arabic poetic understanding in LLMs.

\section{Fann or Flop Data Statistics}

To better characterize the distributional properties of our curated Arabic poetry dataset, we present a series of descriptive statistics that cover both historical and thematic dimensions.

\begin{figure}[hptb]
\centering
\begin{subfigure}[t]{0.5\textwidth}
    \centering
    \includegraphics[width=\textwidth]{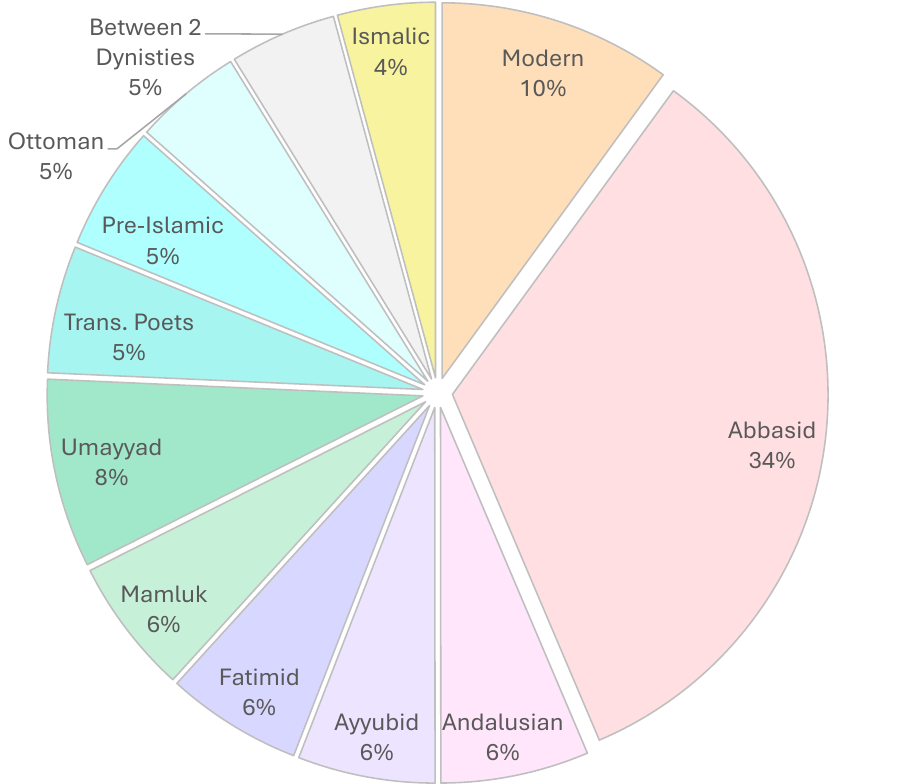}
    \caption{
    \small
    \textbf{Distribution of poems by historical era.} The chart shows the proportion of poems collected from each era. Abbasid, Modern,  and Andalusian periods are the most represented, reflecting their central role in Arabic literary production.
    }
    \label{fig:era_dist}
\end{subfigure}

\vspace{1em}  

\begin{subfigure}[hptb]{0.5\textwidth}
    \centering
    \includegraphics[width=\textwidth]{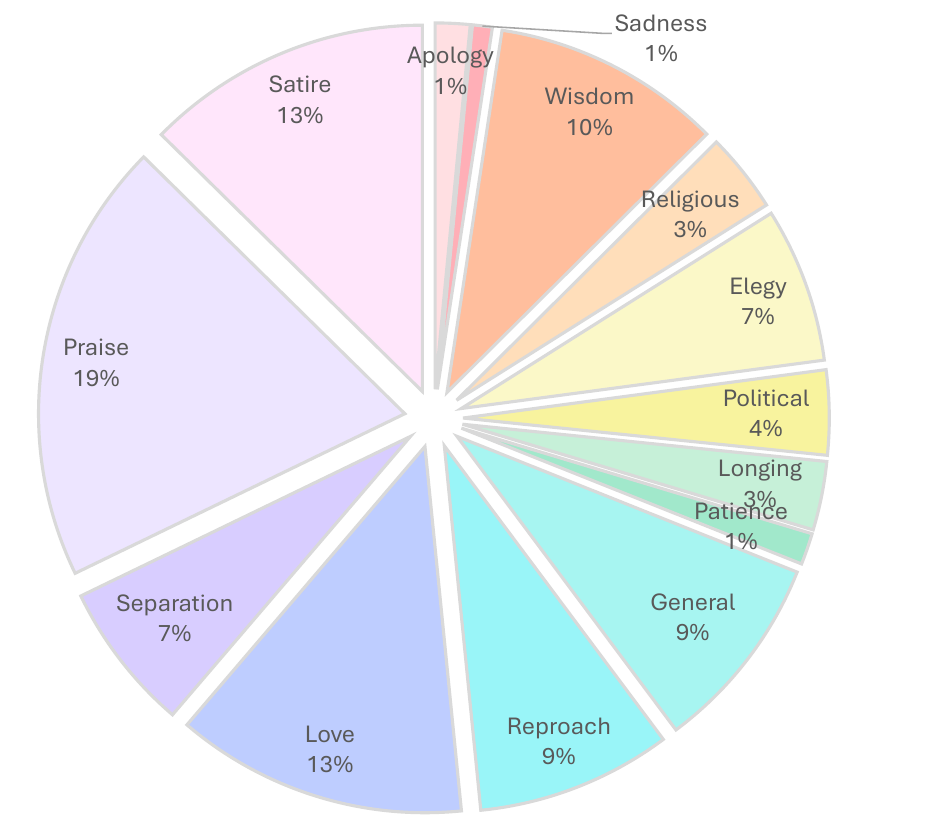}
    \caption{
    \small
    \textbf{Distribution of poems by genre.} This chart shows the proportion of poetic genres across the dataset. Praise, Satire, and Love dominate the distribution, while genres such as Apology and Sadness appear less frequently.
    }
    \label{fig:genre_dist}
\end{subfigure}

\caption{
\small
\textbf{Era and Genre Statistics.} Subfigure~(a) displays the distribution of poems across historical eras, while subfigure~(b) shows the overall genre distribution across the dataset.
}
\label{fig:era_genre_overview}
\end{figure}

\begin{figure*}[t!]
\centering  
\includegraphics[width=\textwidth]{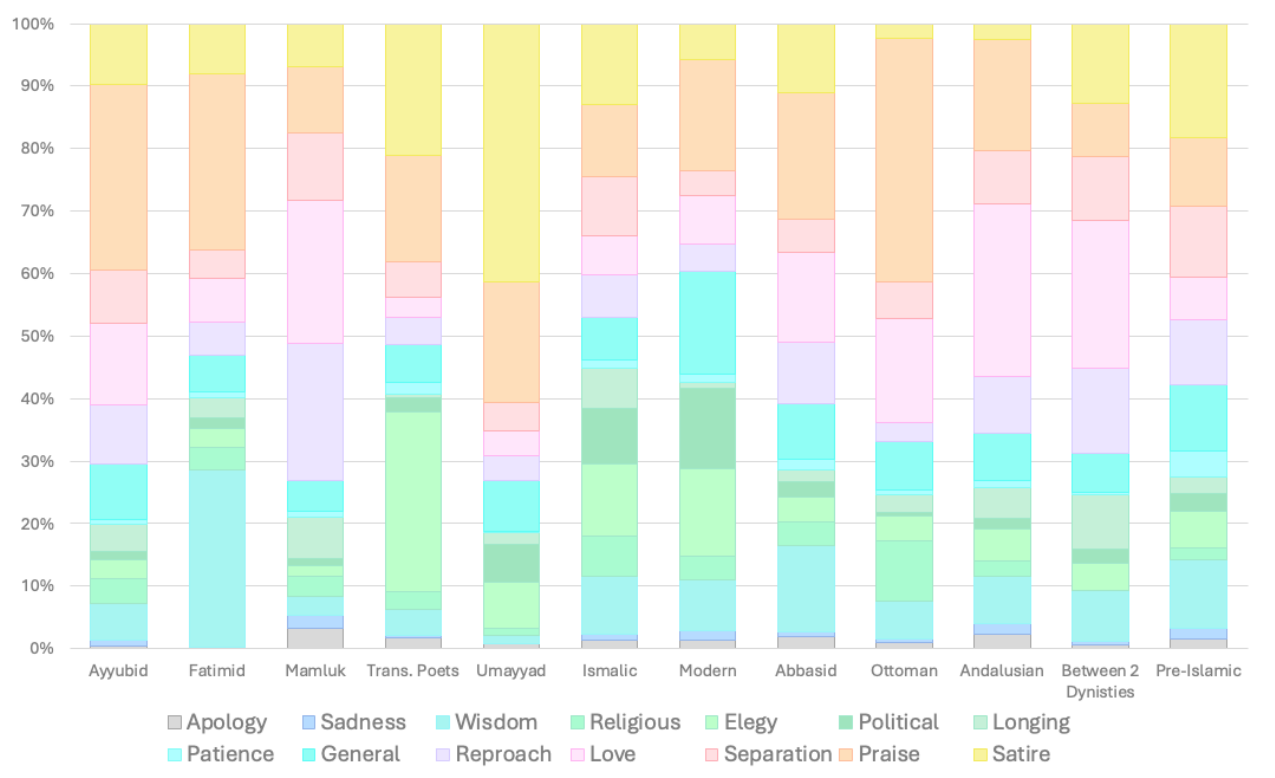}
\caption{
\small
\textbf{Genre distribution across historical eras.} This stacked bar chart illustrates how poetic themes evolved across different dynasties. It highlights patterns such as the prominence of Praise and Satire during the Abbasid and Umayyad eras, and the diverse thematic expression in Modern poetry.
}
\label{fig:genre_era}
 \vspace{-1em}
\end{figure*}

These include the distribution of poems across major eras (Figure~\ref{fig:era_dist}), the overall distribution of poetic genres (Figure~\ref{fig:genre_dist}), and a genre-by-era breakdown (Figure~\ref{fig:genre_era}).

Complementing these visualizations, we also include detailed per-era tables listing the most represented poets and the number of poems attributed to each. Together, these statistics contextualize the coverage of the dataset and support downstream applications such as genre classification, diachronic literary analysis, and poet-specific modeling.

Tables~\ref{tab:preislamic_poets} to~\ref{tab:ottoman_poets} provide a breakdown of the number of poems attributed to prominent Arabic poets across different historical eras. Each table is dedicated to one era:\\
Table~\ref{tab:preislamic_poets}: Pre-Islamic era;
Table~\ref{tab:islamic_transitional_poets}
Transitional (Early Islamic) period;
Table~\ref{tab:modern_poets}: Modern era;
Table~\ref{tab:islamic_poets}: Islamic era; 
Table~\ref{tab:umayyad_poets}: Umayyad era;
Table~\ref{tab:abbasid_poets}: Abbasid era;
Table~\ref{tab:btwen_poets}: Between 2 Dynasties;
Table~\ref{tab:fatimid_poets}: Fatimid Dynasty;
Table~\ref{tab:andalusian_poets}: Andalusian era;
Table~\ref{tab:ayyubid_poets}: Ayybid era.
Table~\ref{tab:mamluk_poets}: Mamluk Dynasty;
Table~\ref{tab:ottoman_poets}: Ottoman era.\\
 \vspace{-1em}

\begin{table}[hptb]
\centering
\small
\setlength{\tabcolsep}{17pt}
\resizebox{\columnwidth}{!}{%
\begin{tabular}{l r}
\rowcolor{gray!20}
\textbf{Era} & \textbf{Pre-Islamic} \\
\toprule
\rowcolor{gray!10}
\textbf{Poet} & \textbf{Poems} \\
Aws ibn Hajar & 35 \\
al-Samaw'al & 12 \\
al-Sulayk ibn al-Sulaka & 7 \\
Imru' al-Qais ibn Hujr & 34 \\
Zuhayr ibn Abi Sulma & 48 \\
Salama ibn Jandal & 14 \\
Tarfah ibn al-Abd & 26 \\
Urwah ibn al-Ward al-Absi & 31 \\
Ubayd ibn al-Abras & 40 \\
Amr ibn Qami'a & 21 \\
Amr ibn Kulthum & 24 \\
Antarah ibn Shaddad & 82 \\
\rowcolor{gray!10}
\textbf{Total} & \textbf{374} \\
\bottomrule
\end{tabular}%
}
\vspace{-1em}
\caption{\small Poem counts for major poets from the Pre-Islamic era.}
\label{tab:preislamic_poets}
\end{table}

\begin{table}[hptb]
\centering
\small
\setlength{\tabcolsep}{17pt}
\resizebox{\columnwidth}{!}{%
\begin{tabular}{l r}
\rowcolor{gray!20}
\textbf{Era} & \textbf{Transitional Poet} \\
\toprule
\rowcolor{gray!10}
\textbf{Poet} & \textbf{Poems} \\
Al-Hadira & 7 \\
Al-Hutay'a & 95 \\
Al-Khansa & 92 \\
Hassan ibn Thabit & 74 \\
Amir ibn al-Tufayl & 41 \\
Amr ibn Barraqa & 5 \\
Labid ibn Rabi'a & 69 \\
\rowcolor{gray!10}
\textbf{Total} & \textbf{383} \\
\bottomrule
\end{tabular}%
}
\vspace{-1em}
\caption{\small Poem counts for major poets from the Early-Islamic Transitional period.}
\label{tab:islamic_transitional_poets}
\end{table}
 \vspace{-1em}

\begin{table}[hptb]
\centering
\small
\setlength{\tabcolsep}{22pt}
\resizebox{\columnwidth}{!}{%
\begin{tabular}{l r}
\rowcolor{gray!20}
\textbf{Era} & \textbf{Modern} \\
\toprule
\rowcolor{gray!10}
\textbf{Poet} & \textbf{Poems} \\
\small Ahmed Shawqi &\small 460 \\
\small Hafiz Ibrahim & \small 240 \\
\rowcolor{gray!10}
\small \textbf{Total} & \small \textbf{700} \\
\bottomrule
\end{tabular}%
}   
\vspace{-1em}
\caption{\small Poem counts for major poets from the Modern era.}
\label{tab:modern_poets}
\end{table}

\begin{table}[hptb]
\centering
\small
\setlength{\tabcolsep}{5pt}
\resizebox{\columnwidth}{!}{%
\begin{tabular}{l r}
\rowcolor{gray!20}
\textbf{Era} & \textbf{Islamic} \\
\toprule
\rowcolor{gray!10}
\textbf{Poet} & \textbf{Poems} \\
Abu Muhammad al-Faq'asi & 28 \\
Al-Akraa bin Muath Al-Kushairi & 16 \\
Asmaa Bin Kharja El-Fazari & 4 \\
Aasha Taroud & 4 \\
Khuzaima ibn Thabit al-Ansari & 14 \\
Khalid ibn al-Walid & 7 \\
Az-Zzubayr bin Al-Awam & 4 \\
As-Samhari Al-Okliy & 11 \\
Al-Ghitamish Al-Dabbi & 4 \\
Abd al-Rahman ibn Abi Bakr al-Siddiq & 3 \\
Jubaiha' al-Ashja'i & 7 \\
Habib ibn Khidrah al-Hilali & 6 \\
Ka'b ibn Mashhur al-Makhbali & 13 \\
Mas'ud al-Mazini & 3 \\
Satira al-Usaybiyya & 4 \\
Ziyad ibn Abihi & 3 \\
Ziyad ibn Hanzala al-Tamimi & 4 \\
Murrah ibn Junada & 3 \\
Atika bint Zayd & 6 \\
Abd al-Aziz ibn Zararah al-Kalabi & 6 \\
Urwa ibn Hizam & 6 \\
Ali ibn al-Husayn & 6 \\
Amr ibn al-'As & 26 \\
Amra bint Mirdas & 4 \\
\rowcolor{gray!10}
Other Poet with 1 or 2 poems & 102 \\
\rowcolor{gray!10}
\textbf{Total} & \textbf{294} \\
\bottomrule
\end{tabular}%
}
\vspace{-1em}
\caption{\small Poem counts for major poets from the Islamic era.}
\label{tab:islamic_poets}
\end{table}

\begin{table}[hptb]
\centering
\small
\setlength{\tabcolsep}{8pt}
\resizebox{\columnwidth}{!}{%
\begin{tabular}{l r}
\rowcolor{gray!20}
\textbf{Era} & \textbf{Umayyad Era} \\
\toprule
\rowcolor{gray!10}
\textbf{Poet} & \textbf{Poems} \\
Al-Akhtal & 136 \\
Jarir & 239 \\
Al-Farazdak& 178 \\
Ubaydallah ibn al-Ruqayyat & 15 \\
\rowcolor{gray!10}
\textbf{Total} & \textbf{568} \\
\bottomrule
\end{tabular}%
}
\vspace{-1em}
\caption{\small Poem counts for major poets from the Umayyad Era.}
\label{tab:umayyad_poets}
\end{table}

\vspace{-2em}

\begin{table}[hptb]
\centering
\small
\setlength{\tabcolsep}{17pt}
\resizebox{\columnwidth}{!}{%
\begin{tabular}{l r}
\rowcolor{gray!20}
\textbf{Era} & \textbf{Abbasid} \\
\toprule
\rowcolor{gray!10}
\textbf{Poet} & \textbf{Poems} \\
Abu al-Atahiya & 362 \\
Abu Firas al-Hamdani & 129 \\
Abu Nuwas & 701 \\
Abu Tammam & 29 \\
Ibn al-Rumi & 227 \\
Imam al-Shafi'i & 20 \\
al-Buhturi & 601 \\
al-Mutanabbi & 273 \\
\rowcolor{gray!10}
\textbf{Total} & \textbf{2342} \\
\bottomrule
\end{tabular}%
}
\vspace{-1em}
\caption{\small Poem counts for major poets from the Abbasid era.}
\label{tab:abbasid_poets}
\end{table}

\vspace{-1em}

\begin{table}[hptb]
\centering
\small
\resizebox{\columnwidth}{!}{%
\begin{tabular}{l r}
\rowcolor{gray!20}
\textbf{Era} & \textbf{Between Dynasties} \\
\toprule
\rowcolor{gray!10}
\textbf{Poet} & \textbf{Poems} \\
Bashar bin Bord& 321 \\
\rowcolor{gray!10}
\textbf{Total} & \textbf{321} \\
\bottomrule
\end{tabular}%
}
\vspace{-1em}
\caption{\small Poem counts for major poets from the Between 2 Dynasties.}
\label{tab:btwen_poets}
\end{table}
\vspace{-1em}

\begin{table}[hptb]
\centering
\small
\setlength{\tabcolsep}{8pt}
\resizebox{\columnwidth}{!}{%
\begin{tabular}{l r}
\rowcolor{gray!20}
\textbf{Era} & \textbf{Fatimid Dynasty} \\
\toprule
\rowcolor{gray!10}
\textbf{Poet} & \textbf{Poems} \\
Abu al-Ala al-Ma'arri & 183 \\
Ibn Hayyus & 120 \\
Arqala al-Kalbi & 106 \\
\rowcolor{gray!10}
\textbf{Total} & \textbf{409} \\
\bottomrule
\end{tabular}%
}
\vspace{-1em}
\caption{\small Poem counts for major poets from the Fatimid Dynasty.}
\label{tab:fatimid_poets}
\end{table}

\vspace{-1em}

\begin{table}[hptb]
\centering
\small
\setlength{\tabcolsep}{10pt}
\resizebox{\columnwidth}{!}{%
\begin{tabular}{l r}
\rowcolor{gray!20}
\textbf{Era} & \textbf{Andalusian} \\
\toprule
\rowcolor{gray!10}
\textbf{Poet} & \textbf{Poems} \\
Abu Ishaq al-Albiri & 38 \\
Ibn Khafaja & 225 \\
Ibn Zaydun & 146 \\
Ibn Sahl al-Andalusi & 37 \\
\rowcolor{gray!10}
\textbf{Total} & \textbf{446} \\
\bottomrule
\end{tabular}%
}
\vspace{-1em}
\caption{\small Poem counts for major poets from the Andalusian era.}
\label{tab:andalusian_poets}
\end{table}

\begin{table}[hptb]
\centering
\small
\resizebox{\columnwidth}{!}{%
\begin{tabular}{l r}
\rowcolor{gray!20}
\textbf{Era} & \textbf{Ayyubid Dynasty} \\
\toprule
\rowcolor{gray!10}
\textbf{Poet} & \textbf{Poems} \\
Ibn al-Farid & 35 \\
Sibt Ibn al-Tawawidhi & 291 \\
Muhyiddin Ibn Arabi & 87 \\
\rowcolor{gray!10}
\textbf{Total} & \textbf{413} \\
\bottomrule
\end{tabular}%
}
\vspace{-1em}
\caption{\small Poem counts for major poets from the Ayyubid Dynasty.}
\label{tab:ayyubid_poets}
\end{table}

\vspace{-1em}

\begin{table}[hptb]
\centering
\small
\resizebox{\columnwidth}{!}{%
\begin{tabular}{l r}
\rowcolor{gray!20}
\textbf{Era} & \textbf{Mamluk Dynasty} \\
\toprule
\rowcolor{gray!10}
\textbf{Poet} & \textbf{Poems} \\
Baha al-Din Zuhayr & 368 \\
Safiyy al-Din al-Hilli & 40 \\
\rowcolor{gray!10}
\textbf{Total} & \textbf{408} \\
\bottomrule
\end{tabular}%
}
\vspace{-1em}
\caption{\small Poem counts for major poets from the Mamluk Dynasty.}
\label{tab:mamluk_poets}
\end{table}

\begin{table}[hptb]
\centering
\small
\resizebox{\columnwidth}{!}{%
\begin{tabular}{l r}
\rowcolor{gray!20}
\textbf{Era} & \textbf{Ottoman} \\
\toprule
\rowcolor{gray!10}
\textbf{Poet} & \textbf{Poems} \\
Abu al-Ma'ali al-Talawi & 75 \\
Ibn Razka & 19 \\
Ibn Matuq al-Musawi & 74 \\
al-Kawkabani & 2 \\
Bint al-Shuhna & 2 \\
Abd al-Rahman al-Musili & 58 \\
Muhammad al-Isba'i & 31 \\
Muhammad al-Sharafi al-Safaqsi & 65 \\
\rowcolor{gray!10}
\textbf{Total} & \textbf{326} \\
\bottomrule
\end{tabular}%
}
\vspace{-1em}
\caption{\small Poem counts for major poets from the Ottoman era.}
\label{tab:ottoman_poets}
\end{table}
\vspace{-1em}

\begin{figure}[t!]
\centering  
\includegraphics[width=0.48\textwidth, height=20.5cm]{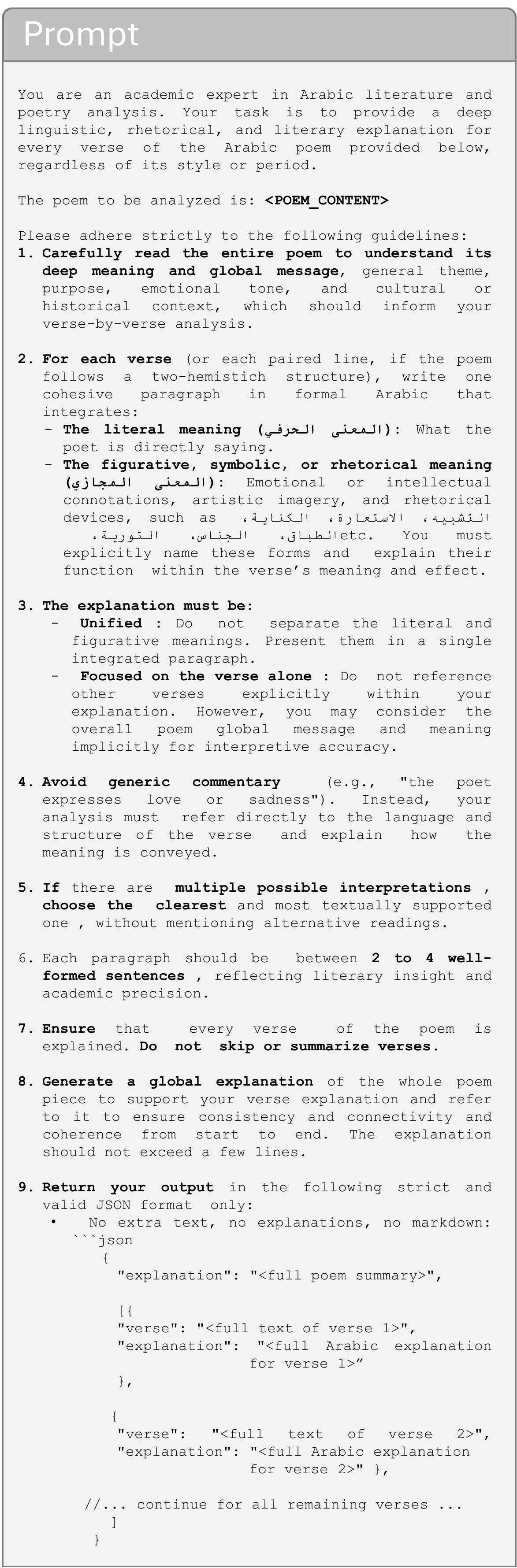}
\caption{
\small
\textbf{The verse-level explanation prompt used for evaluation.} This prompt instructs the model to produce detailed verse-by-verse explanations in Arabic. It guides the model to integrate both literal and figurative interpretations, explicitly name rhetorical devices (e.g., metaphor, personification, paronomasia). The prompt enforces coherence, academic rigor, and structural consistency by requiring output in a strict JSON format.
}
\label{fig:prompt}
 \vspace{-1em}
\end{figure}
\newpage

\section{Prompts Used}

\subsection{Model Generation Prompt}

To generate verse-level explanations suitable for evaluating both open- and closed-source models, we developed a carefully optimized generation prompt. The prompt design followed an iterative and augmented process. Initially, we used a simple bilingual (in Arabic and English) prompt asking for explanations. Based on early outputs, which tended to capture local semantic meaning but lacked coherence and global context, we progressively refined the prompt to elicit more structured and connected responses.

\begin{figure*}

\begin{tcolorbox}[colback=gray!5, colframe=black!45,
    fonttitle=\bfseries, title=System Prompt used to evaluate the poem explanation for Faithfulness and Fluency metric
    ]
\small

\lstset{breaklines=true, basicstyle=\small\ttfamily} 

\begin{lstlisting}
You are an expert Arabic linguist and literary evaluator.

Your task is to **evaluate a full Arabic poem's verse-by-verse explanations**.
You will compare **ground-truth** (human-written) explanations with 
**generated** explanations from an AI model.

You will judge each verse explanation based on the following two criteria:

---

### Evaluation Criteria (per verse)

1. **Faithfulness / Consistency**:  
   Is the generated explanation consistent with the meaning of the verse?  
   - Score 5: Deeply faithful to the verse's content  
   - Score 3: General alignment but loses poetic imagery  
   - Score 1: Misinterprets or invents meaning

2. **Fluency / Grammaticality**:  
   Is the generated explanation well-formed Modern Standard Arabic?  
   - Score 5: Fluent, grammatically correct  
   - Score 3: Understandable with minor issues  
   - Score 1: Awkward, incomplete, or ungrammatical

---

### What You Will Receive

You will receive for each poem:
- `poem_title`
- "ground_truth": a list of objects { "v": <int>, "text": <string> }
- "generated": a list of objects with the **same v indices**

---

### What You Must Do

- Compare all verses together and assign a single score of 1-5 for each criterion.
- Do **not** provide per-verse scores or any comments.


Then:
- Calculate average scores for the whole poem
- Provide an `overall_score` (1-5) that reflects your judgment across all verses

---

Do NOT provide any comments or rationale.
Respond with valid JSON **only** in this format:

### Output Format (in JSON)

{
  "faithfulness_score": <1-5>,
  "fluency_score": <1-5>,
  "overall_score": <1-5>
}
\end{lstlisting}

\end{tcolorbox}
\caption{
System prompt used for LLM-Judge evaluation of verse-by-verse poem explanations. LLM~\cite{openai2024gpt4ocard} compare AI-generated outputs with original explanations and assign overall scores for faithfulness to meaning and fluency in MSA, following clearly defined criteria. The structured format ensures consistency and reliability across evaluations.
}
\label{fig:eval_prompt}
\end{figure*}

Through multiple rounds of testing, expert evaluation, and prompt engineering, we incorporated explicit instructions to address both local (verse-specific) and global (poem-wide) interpretive elements as support. This enhancement significantly improved the quality of the generated explanations, resulting in outputs that were more coherent, context-aware, and semantically aligned with the original verses.

After extensive comparison, expert reviewers favored the English version of the prompt over its Arabic counterpart, as it more consistently achieved local-global alignment and produced well-connected, high-quality explanations. This final version of the English prompt (Figure~\ref{fig:prompt}) was adopted for all subsequent evaluations.

\subsection{Model Evaluation Prompt}

To ensure consistent and reliable automatic LLM-Judge evaluation of model-generated poem explanations, we designed a clear and structured system prompt (see Figure~\ref{fig:eval_prompt}). The prompt positions the evaluator as an expert Arabic linguist and literary critic, responsible for assessing AI-generated verse-by-verse explanations against ground-truth references.

Each poem is evaluated on two key dimensions: Faithfulness/Consistency, which measures how accurately the explanation reflects the verse's intended meaning, and Fluency/Grammaticality, which assesses the quality of the generated text in Modern Standard Arabic. Annotators assign a score from 1 to 5 for each criterion based on the overall performance across all verses, without providing per-verse feedback or open-ended commentary.

The prompt ensures simplicity, objectivity, and high inter-annotator agreement, making it well-suited for evaluating poetic reasoning in culturally rich and linguistically nuanced contexts like Arabic poetry.

 \begin{table*}[t!]
\centering
\small
\setlength{\tabcolsep}{6pt}
\renewcommand{\arraystretch}{1.15}
\resizebox{\textwidth}{!}{%
\begin{tabular}{llcccccc}
\toprule
\rowcolor{gray!10} & \textbf{Model} & \textbf{Pre-Islamic} & \textbf{Transitional} & \textbf{Early Islamic} & \textbf{Umayyad} & \textbf{Abbasid} & \textbf{Fatimid} \\
\midrule
\multirow{6.2}{*}{\rotatebox{90}{\small{Closed}}} 
 & GPT-4o-2024-08-06~\cite{openai2024gpt4ocard} & 0.6425 & 0.6502 & 0.7116 & 0.6166 & 0.6699 & 0.7050 \\
    & GPT-4o-mini-2024-07-18~\cite{gpt4omini} & 0.4355 & 0.4789 & 0.5436 & 0.4200 & 0.4266 & 0.4532  \\
    & Gemini-2.5-Flash~\cite{gemini2.5flash} & 0.7275 & 0.7308 & 0.7527 & 0.7112 & 0.7417 & 0.7542 \\
    & Gemini-2.0-Flash~\cite{gemini2.0flash} & 0.6908 & 0.7156 & 0.7458 & 0.6798 & 0.7033 & 0.7462 \\
    & Gemini-1.5-Pro~\cite{gemini1.5} & 0.6004 & 0.6372 & 0.6497 & 0.6312 & 0.6035 & 0.6502 \\
    & Fanar-Star~\cite{fanar} & 0.6142 & 0.6354 & 0.6621 & 0.5900 & 0.6413 & 0.6717 \\
\midrule
\multirow{9.2}{*}{\rotatebox{90}{\small{Open}}}
    & Deepseek-V3~\cite{liu2024deepseek} & 0.5066 & 0.5875 & 0.6174 & 0.5482 & 0.4736 & 0.5581 \\
    & Deepseek-R1~\cite{guo2025deepseek} & 0.5066 & 0.5875 & 0.6174 & 0.5482 & 0.4736 & 0.5581 \\
    & Llama-3.3-70B~\cite{llama3.3} & 0.5456 & 0.5469 & 0.5747 & 0.5211 & 0.5341 & 0.5387 \\
    & Qwen-3~\cite{qwen3} & 0.6142 & 0.6354 & 0.6621 & 0.5900 & 0.6413 & 0.6717 \\
    & Aya-Expanse~\cite{dang2024aya} & 0.6142 & 0.6354 & 0.6621 & 0.5900 & 0.6413 & 0.6717 \\
    & ALLaM-7B~\cite{allam} & 0.5619 & 0.5630 & 0.6037 & 0.5844 & 0.5848 & 0.6158 \\
    & Jais~\cite{sengupta2023jais} & 0.6124 & 0.6289 & 0.6482 & 0.6047 & 0.6295 & 0.6421 \\
    & AceGPT-v2-70B-Chat~\cite{huang2023acegpt} & 0.5851 & 0.5656 & 0.6104 & 0.5770 & 0.6119 & 0.6095  \\
\bottomrule
\end{tabular}}

\vspace{1em}

\resizebox{\textwidth}{!}{%
\begin{tabular}{llcccccc}
\toprule
\rowcolor{gray!10} & \textbf{Model} & \textbf{Andalusian} & \textbf{Ayyubid} & \textbf{Mamluk} & \textbf{Between Dynasties} & \textbf{Ottoman} & \textbf{Modern } \\
\midrule
\multirow{4.2}{*}{\rotatebox{90}{\small{Closed}}} 
& GPT-4o-2024-08-06~\cite{openai2024gpt4ocard} & 0.7128 & 0.6774 & 0.7393 & 0.6656 & 0.7379 & 0.6843 \\
    & GPT-4o-mini-2024-07-18~\cite{gpt4omini} & 0.4869 & 0.4303 & 0.4507 & 0.4240 & 0.4836 & 0.3988  \\
    & Gemini-2.5-Flash~\cite{gemini2.5flash} & 0.7778 & 0.7416 & 0.7866 & 0.7398 & 0.7994 & 0.7544 \\
    & Gemini-2.0-Flash~\cite{gemini2.0flash} & 0.7527 & 0.7320 & 0.7698 & 0.7164 & 0.7585 & 0.6951 \\
    & Gemini-1.5-Pro~\cite{gemini1.5} & 0.6710 & 0.6074 & 0.6377 & 0.5971 & 0.6441 & 0.5965 \\
    & Fanar-Star~\cite{fanar} & 0.6749 & 0.6454 & 0.7105 & 0.6342 & 0.7151 & 0.6429 \\
\midrule
\multirow{9.2}{*}{\rotatebox{90}{\small{Open}}}
    & Deepseek-V3~\cite{liu2024deepseek} & 0.5927 & 0.5065 & 0.5448 & 0.4929 & 0.5226 & 0.4705 \\
    & Deepseek-R1~\cite{guo2025deepseek} & 0.5927 & 0.5065 & 0.5448 & 0.4929 & 0.5226 & 0.4705 \\
    & Llama-3.3-70B~\cite{llama3.3} & 0.5873 & 0.5221 & 0.5849 & 0.5129 & 0.5712 & 0.4897 \\
    & Qwen-3~\cite{qwen3} & 0.6749 & 0.6454 & 0.7105 & 0.6342 & 0.7151 & 0.6429 \\
    & Aya-Expanse~\cite{dang2024aya} & 0.6749 & 0.6454 & 0.7105 & 0.6342 & 0.7151 & 0.6429 \\
    & ALLaM-7B~\cite{allam} & 0.5892 & 0.6044 & 0.6736 & 0.5905 & 0.6556 & 0.6302 \\
    & Jais~\cite{sengupta2023jais} & 0.6540 & 0.6399 & 0.6812 & 0.6183 & 0.6625 & 0.6348 \\
    & AceGPT-v2-70B-Chat~\cite{huang2023acegpt} & 0.6215 & 0.6131 & 0.6683 & 0.5681 & 0.6273 & 0.6044  \\
\bottomrule
\end{tabular}}

\vspace{-0.5em}
\caption{
\textbf{Era-wise Evaluation using Textual Entailment (TE).} Era-wise performance of closed and open-source models on the Arabic poem understanding task, measured using the Textual Entailment metric. This metric evaluates how well the model-generated explanation logically aligns with the original poem content. The results are grouped across key historical eras, from Pre-Islamic to Modern, allowing a fine-grained view of model strengths and limitations across time periods. Closed models such as GPT-4o and Gemini variants demonstrate consistently high entailment across most eras, while select open models like Deepseek-V3 and Aya-Expanse show promising results in specific historical contexts. This analysis highlights the importance of temporal generalization and cultural grounding in building robust Arabic literary reasoning models.
}
\label{tab:performance_era_add}
\end{table*}

\begin{figure*}[t]
\centering  
\includegraphics[width=\textwidth]{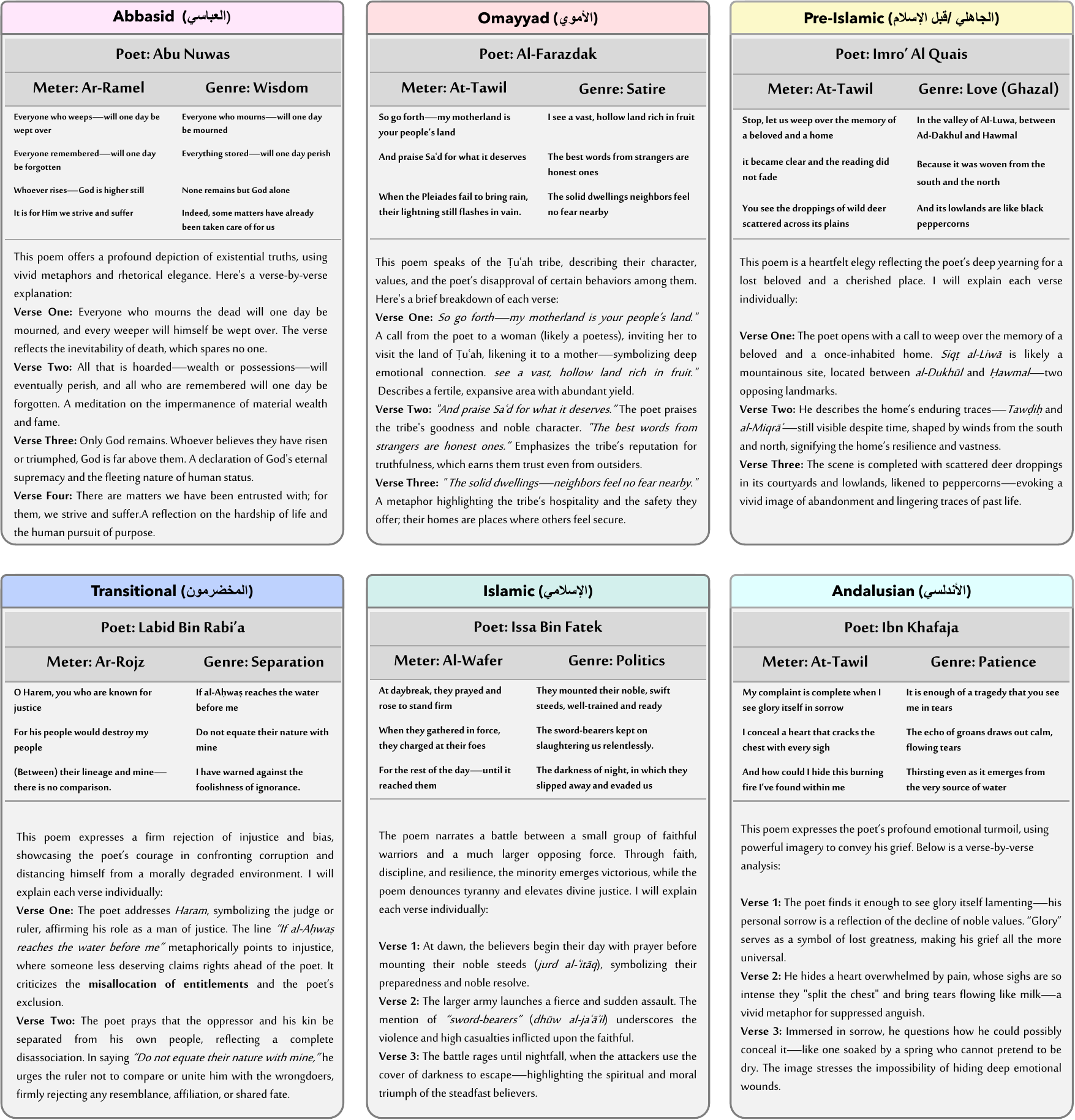}
\vspace{-1.5em}
\caption{
\small
\textbf{Translated Samples.} This figure presents English translations of the Arabic samples shown in Figure~\ref{fig:dataset_sample}. The translations are included to facilitate understanding and accessibility for non-Arabic speakers, allowing broader engagement with the poetic content without requiring prior knowledge of Arabic.}
\label{fig:era_sample_en}
\vspace{-1em}
\end{figure*}

\section{Additional Examples: Qualitative, Translated, and Quantitative Insights}
\label{app_additional}
In the following section, we present a more detailed evaluation of the Textual-Entailment (Refer Table~\ref{tab:performance_era_add})  metric across the 12 historical eras, comparing both open-source and closed-source models on this dimension. To support comprehensive engagement and a clearer understanding of the data evaluated, we also include selected English translations of Arabic poetic samples as well as additional qualitative Arabic examples. These additions offer deeper insight into the linguistic diversity, thematic range, and overall quality of the dataset used in our analysis.

\subsection{English Translated Qualitative Samples}
\label{app_en_trans}
To support accessibility and improve cross-linguistic understanding, we translated selected Arabic poetry samples shown in Figure~\ref{fig:dataset_sample} into English. For this task, we used GPT-4o, which provided deeper context-aware translations that more accurately capture the figurative and rhetorical nuances of the original verse, outperforming basic literal tools such as Google Translate. These translations allow non-Arabic speakers to more easily engage with the literary richness, emotional depth, and stylistic variety explored in our research.

\subsection{Additional Qualitative Samples}
\label{app_eadd_samples}
\begin{figure*}[t]
\centering  
\includegraphics[width=\textwidth, height=22cm]{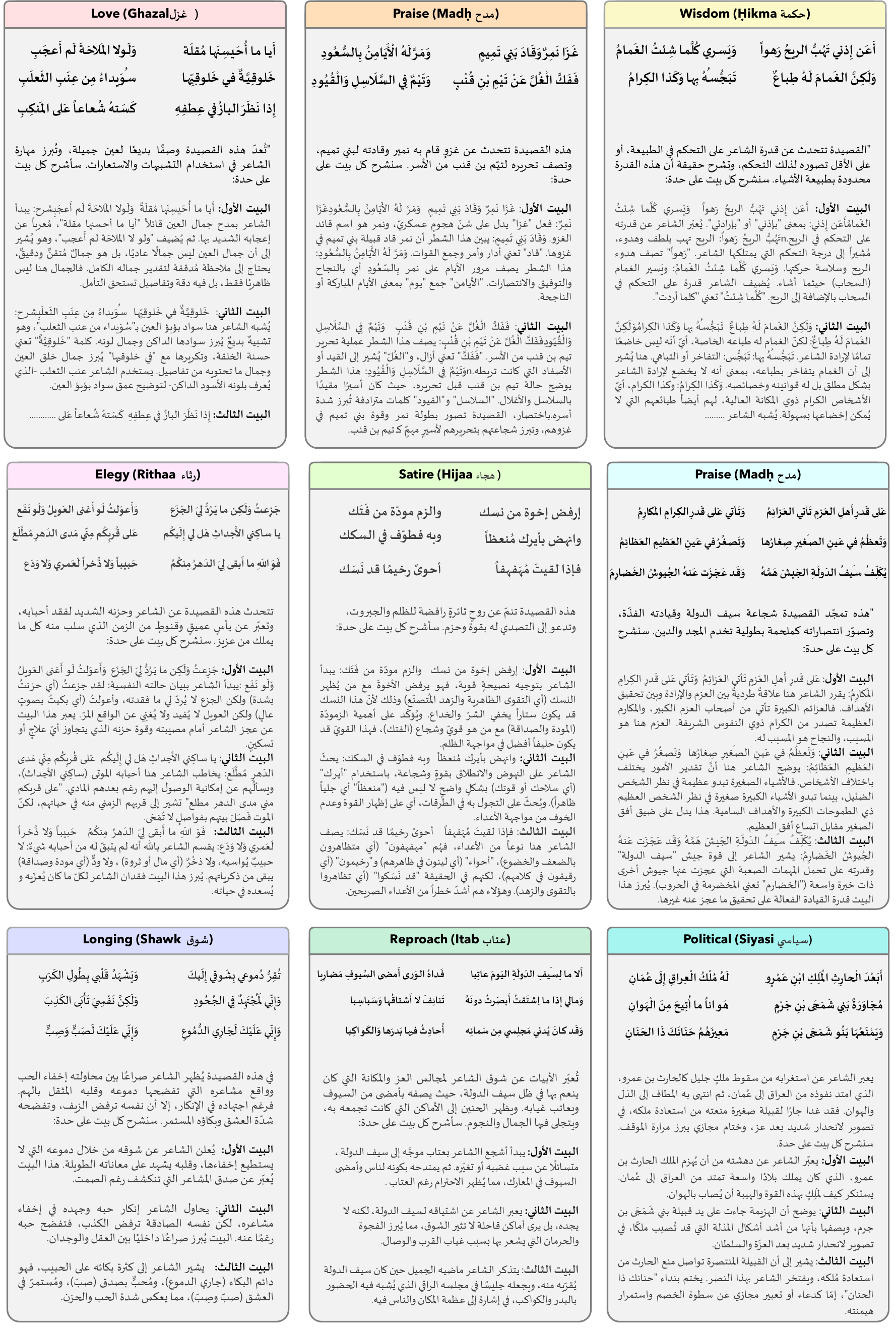}
\vspace{-2em}
\caption{
\small
\textbf{Fann or Flop Samples by Genre.} Additional representative examples from the Fann or Flop benchmark, illustrating the diversity of genres covered, including Love (Ghazal), Praise (Madḥ), Wisdom (Hikma), Satire (Hijā’), Elegy (Rithā’), Reproach ('Itāb), Political Poetry, and Longing (Shawq). Each example showcases a poetic excerpt alongside an interpretive breakdown highlighting figurative language, rhetorical devices, and thematic nuances. These curated samples reflect the benchmark’s aim to evaluate models’ nuanced understanding of Arabic poetic tradition.
}
\label{fig:additional_genre_era_sample}

 \vspace{-1em}
\end{figure*}

To further showcase the dataset's richness, we present additional qualitative samples spanning diverse historical periods, poetic genres (e.g., satire, elegy, political verse), and metrical patterns (See Figure ~\ref{fig:additional_genre_era_sample}). These examples were selected to demonstrate the stylistic, thematic, and rhetorical variety encountered in our evaluation.

\end{document}